\documentclass[11pt]{article}

\usepackage[final]{acl}

\usepackage{times}
\usepackage{latexsym}
\usepackage{amsmath}
\usepackage{tabularx}
\usepackage[T1]{fontenc}

\usepackage[utf8]{inputenc}

\usepackage{microtype}

\usepackage{inconsolata}

\usepackage{graphicx}
\usepackage{booktabs}
\usepackage{multirow}
\usepackage{amssymb}
\usepackage[table]{xcolor}
\title{\textsc{Globe}: Evolving Structured Agentic Memory for E-commerce Product Optimization in Generative Search}

\author{Hyunseo Kim, Suil Choe, Sukkyoung So, Daeyeon Yi\\
        AI Lab, CTO Division, LG Electronics, South Korea\\
        \{hyunseo00.kim, suil.choe, sukkyoung.so, daeyeon.yi\}@lge.com}
\usepackage{booktabs}
\usepackage{pgfplots}
\begin{document}
\maketitle
\begin{abstract}

Optimizing product listings for generative search requires language agents not only to rewrite effectively but also to continually self-improve by retaining successful strategies across diverse categories. 
However, existing approaches often treat structured listings as free-form text, leaving effective strategies implicit in prompts, search trajectories, or rewritten outputs rather than accumulating them as a reusable agentic memory. 
To address this limitation, we introduce \textsc{Globe}, an agentic framework that drives continual self-improvement by storing optimization knowledge in an evolving handbook of explicit, field-scoped behaviors. 
In each episode, an Editor applies relevant behaviors and receives rank-based feedback, which a Reflector converts into behavior updates. 
Crucially for dependable execution in real-world settings, a Moderator removes redundant proposals, a Curator commits accepted changes, and consistently harmful behaviors are deactivated. 
This cycle transforms episodic feedback into a persistent, field-scoped optimization memory.
Experiments show that structured behaviors outperform monolithic prompt optimization, while feedback-driven evolution and moderation further improve effectiveness and reduce downside risk.
The learned behaviors also remain effective on unseen product categories, suggesting that agentic memory evolution supports interpretable and transferable product optimization.
\end{abstract}

\section{Introduction}
\begin{figure*}[t]
    \centering
    \includegraphics[width=\textwidth]{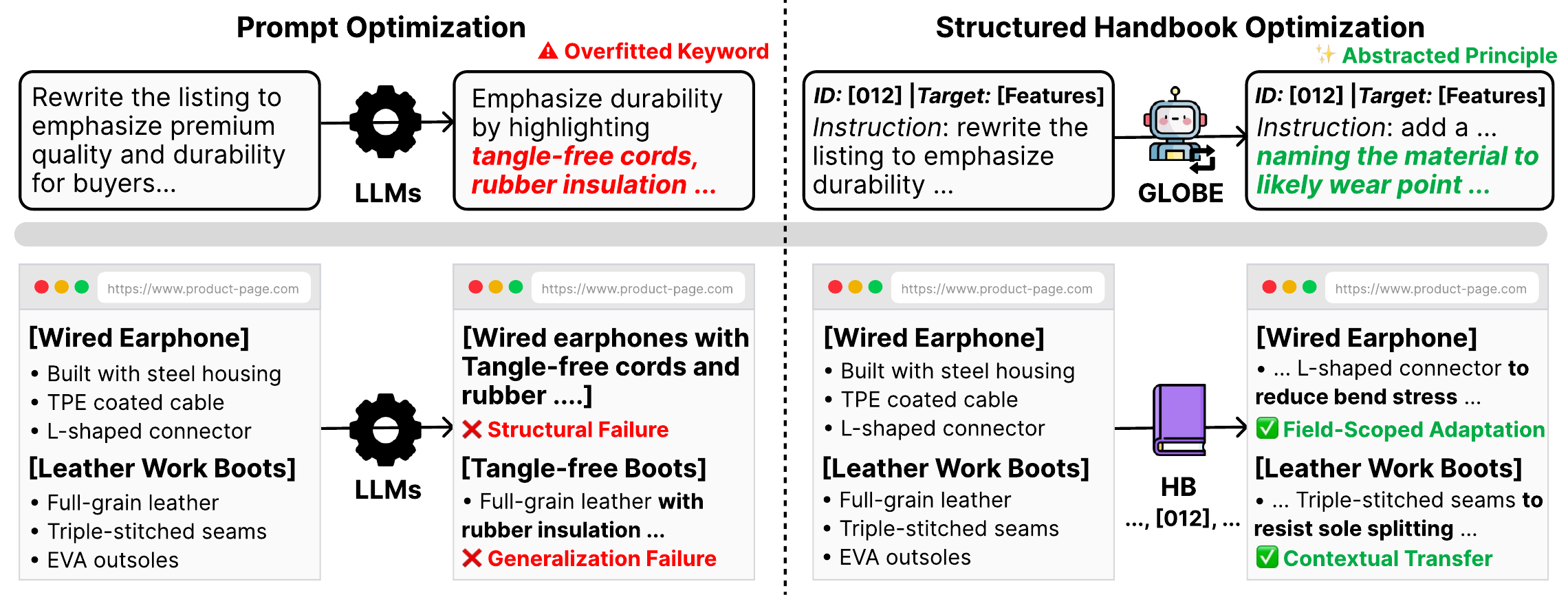}
    \caption{Comparison between monolithic prompt optimization and \textsc{Globe}'s structured handbook optimization. A generic rewrite prompt performs an undifferentiated text transformation, whereas each \textsc{Globe} behavior specifies both a concrete editing operation and its permitted target field, enabling constrained reuse across products.}
    \label{fig:motivation}
\end{figure*}

Language agents increasingly improve themselves not by updating weights, but by accumulating experience in an external memory that is continually refined through feedback~\citep{shinn2023reflexion,wang2023voyager}. 
Building on these advances, agent-centric frameworks are expanding into e-commerce, as large language models (LLMs) shift search toward generative engines that directly compare and rank products by reasoning over rich textual contexts~\citep{li-etal-2024-large,sun-etal-2023-chatgpt,hou2024large,10.1145/3726302.3730055}.
In this environment, an agent optimizing a product listing must shift from conventional search engine optimization methods~\citep{davis2006search,10.1109/MC.2015.297} to strategic content structuring, as its success heavily depends on how favorably the downstream generative engine interprets the presented information.

A key challenge in this real-world setting lies in the structured nature of e-commerce product data. 
Product listings are not arbitrary free-form documents but consist of distinct fields, such as titles, feature bullets, and descriptions, each serving a different functional role.
Nevertheless, prior methods commonly optimize a listing through a single prompt that rewrites the document as a whole \citep{bagga2025egeo,yuan2026agenticgeo}. 
This monolithic approach lacks structural boundaries, treating optimization as unstructured text generation. 
As illustrated in Figure~\ref{fig:motivation}, relying on generic prompt optimization leads to critical structural errors, because the agent cannot restrict specific editing operations to their appropriate target fields.

Beyond this lack of structural control, current approaches fail to generalize successful strategies across multiple optimization episodes.
Even if a specific rewrite tactic successfully improves a product's rank in one category, monolithic prompt optimization cannot abstract this success into a reusable principle. 
As further demonstrated in Figure~\ref{fig:motivation}, when a generic prompt attempts to transfer a durability strategy to new categories, it fails to adapt the context, forcing overfitted keywords (e.g., applying "rubber insulation" to leather boots) rather than grounding the edit in existing product facts.
Consequently, effective strategies must be repeatedly rediscovered, making optimization dependent on fragmented and costly trial and error. 
Product optimization therefore requires an explicit agentic memory mechanism that transforms transient ranking outcomes into persistent, generalizable editing knowledge~\citep{shinn2023reflexion,zhao2024expel}.

To address these limitations, we introduce \textbf{GLOBE} (\textbf{G}enera\textbf{L}izable \textbf{O}ptimization \textbf{B}ehavior \textbf{E}volution), an agentic framework that represents product optimization knowledge as a structured behavior handbook. 
Rather than relying on a single unconstrained rewrite instruction, \textsc{Globe} decomposes optimization into explicit and reusable editing \textit{behaviors}. 
Each behavior specifies a concrete editing operation, the listing fields in which the operation may be applied, and metadata summarizing its historical ranking
outcomes. 
For example, a behavior may instruct the agent to specify the material of a likely wear point, strictly limiting this addition to the feature field.
This formulation separates \textit{what} should be changed from \textit{where} the change is permitted, providing a more interpretable and structurally bounded interface for product-listing optimization.

\textsc{Globe} further transforms transient ranking feedback into persistent optimization knowledge through an iterative \textit{deployment-and-evolution} cycle. 
The current handbook is repeatedly applied to batches of query-product pairs, producing episode-level feedback from the resulting rank changes. 
Recurring success and failure patterns are then used to refine the handbook, after which the updated behaviors are redeployed in subsequent batches. By repeatedly alternating between handbook deployment and feedback-driven refinement, \textsc{Globe} converts episode-specific ranking outcomes into persistent and reusable optimization knowledge.

This formulation provides two complementary advantages. 
First, field-scoped behaviors explicitly specify both the editing operation
and the product fields in which it may be applied, making individual
strategies more structured and inspectable. 
Second, the evolving handbook preserves and reuses knowledge acquired from past ranking
episodes rather than discarding it after each rewrite. 
We therefore investigate whether field-scoped behavior representations improve ranking effectiveness, whether feedback-driven handbook evolution reduces harmful outcomes, and whether the learned optimization knowledge transfers to unseen product categories.

Our main contributions are as follows:
\begin{itemize}
    \item We propose \textsc{Globe}, an agentic product-optimization
    framework that maintains reusable optimization knowledge as a
    structured memory of field-scoped behaviors.
    
    \item We introduce a feedback-driven evolution mechanism that
    converts episode-level ranking outcomes into persistent editing
    knowledge, while filtering redundant and consistently harmful
    behaviors.
    
    \item We demonstrate that \textsc{Globe} improves ranking
    effectiveness and stability, with learned behaviors transferring
    to unseen product categories and ranking models.
\end{itemize}


\section{Related Work}
\subsection{Generative Engine Optimization for E-commerce}
\label{subsec:ranking_optimization}
Generative Engine Optimization (GEO) aims to improve the visibility of target items in generative search by modifying their textual representations.
Initial approaches showed that strategically injected text sequences can influence item exposure and output rankings~\citep{kumar2024manipulating,jin2026core}.
Subsequent prompt-based methods formulate GEO as document rewriting, using LLMs to modify content according to the inferred preferences of a target generative engine~\citep{bagga2025egeo,wu2025autogeo}.
Recent agentic frameworks further automate this process by iteratively searching for effective document-modification strategies~\citep{yuan2026agenticgeo}.
Despite this progression, existing approaches largely optimize documents through free-form, document-level generation, offering limited control over individual fields of structured product listings.
Moreover, effective strategies typically remain embedded in prompts, search trajectories, or rewritten outputs rather than being retained as explicit and reusable knowledge.

\subsection{Feedback-Driven Knowledge Evolution}
\label{subsec:feedback_driven}
Effective optimization requires systems to not only generate high-quality outputs but also learn from past successes and failures to avoid repeatedly rediscovering the same strategies. 
Driven by this need, a growing line of research studies how feedback can be transformed from a transient correction signal into reusable knowledge~\citep{kim2025bespoke}.
Early self-refinement methods use self- or peer-generated critiques to revise the current output~\citep{madaan2023self,schick2022peer}.
Cross-episode approaches go beyond local revision by abstracting successful and failed trajectories into reusable reflections or lessons~\citep{shinn2023reflexion,zhao2024expel}.
At the system level, automatic prompt optimizers use natural-language critiques, textual feedback, or trajectory-level reflection to revise the instructions governing future executions~\citep{pryzant-etal-2023-automatic, yuksekgonul2024textgrad,agrawal2026gepareflectivepromptevolution}.
Recent memory-centric systems further externalize such abstractions into persistent handbooks, evolving playbooks, or learned memory architectures~\citep{didolkar2025metacognitive,zhang2025agentic,xiong2026learning}.
Together, these works demonstrate a progression from local output correction to persistent knowledge evolution, but largely focus on open-ended reasoning, coding, or planning.
In contrast, structured product optimization requires reusable knowledge to be field-scoped and learned from indirect, relative ranking feedback.

\section{Proposed Method: \textsc{Globe}}

\begin{figure*}[t]
    \centering
    \includegraphics[width=\textwidth]{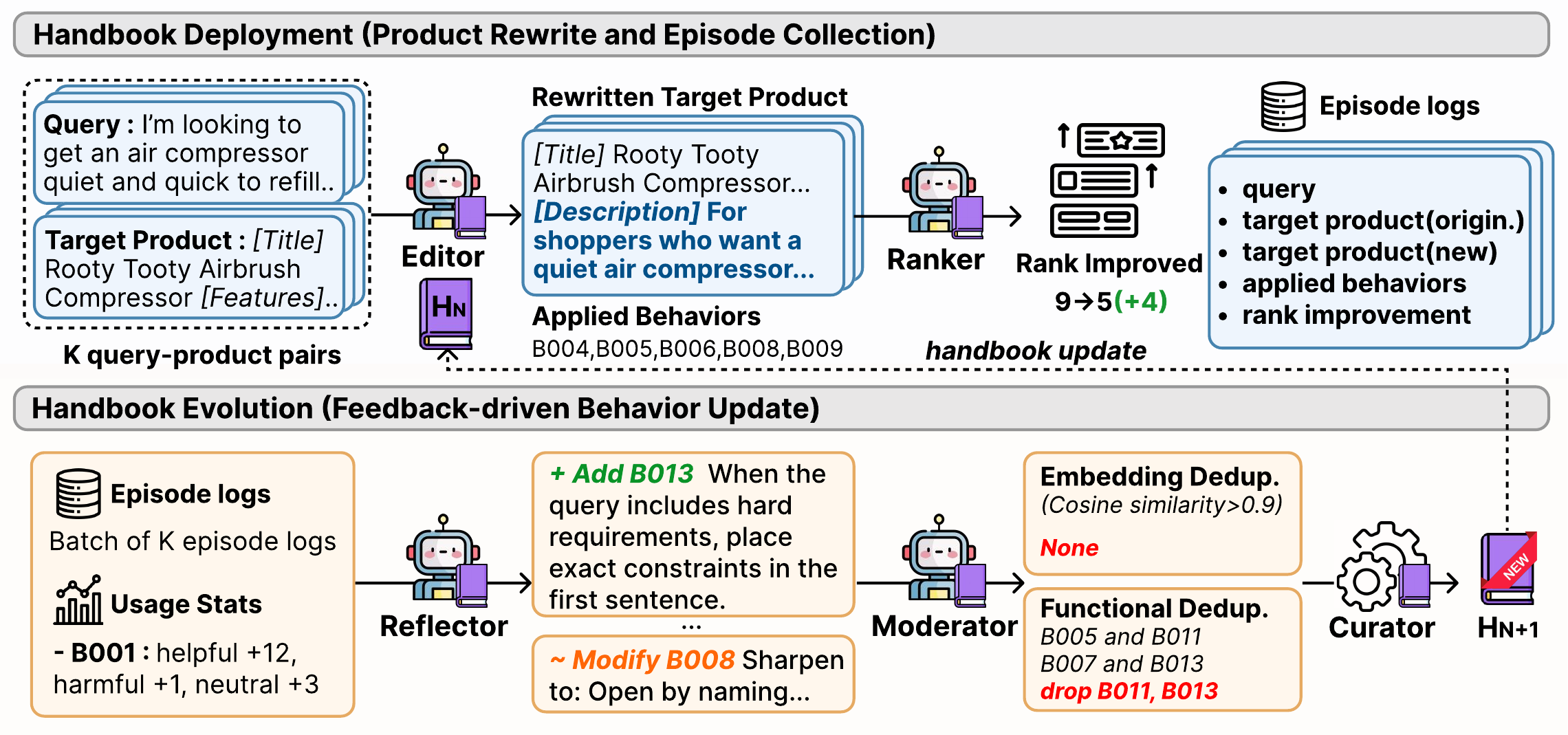}
    \caption{Overview of the \textsc{Globe} framework. Each iteration deploys the active handbook $\mathcal{H}_n$ over a batch of $K$ query-product pairs: the editor rewrites the target listing and the ranker returns episode-level rank feedback. The Reflector then proposes behavior additions or modifications, the Moderator removes redundant proposals, and the Curator commits the survivors into $\mathcal{H}_{n+1}$, which replaces $\mathcal{H}_n$ as the handbook deployed over the next batch. Repeating this \textit{deployment-and-evolution} cycle accumulates episodic feedback into persistent editing knowledge.}
    \label{fig:framework_overview}
\end{figure*}
\textsc{Globe} maintains a structured agentic memory of reusable product-optimization behaviors through an iterative \textit{deployment-and-evolution} cycle, illustrated in Figure~\ref{fig:framework_overview}. 
At iteration $n$, the current
handbook $\mathcal{H}_n$ is deployed across a batch of $K$
query-product pairs. Each deployment produces an episode that records
the rewritten listing, the behaviors applied by the editor, and the
resulting rank change. After the batch is completed, \textsc{Globe}
aggregates these episodes to refine the handbook, producing
$\mathcal{H}_{n+1}$. 
The updated handbook is then deployed in the next iteration, allowing optimization knowledge from
individual ranking outcomes to accumulate as explicit, reusable editing behaviors.

\paragraph{Task Formulation.}
\label{subsec:problem_formulation}
Given a user query $q$ and a candidate product set
$\mathcal{P}=\{p_1,\ldots,p_m\}$, a generative ranking model
$\mathcal{M}$ orders the products according to their relevance to
$q$. We denote the rank of product $p$ by
$r_{\mathcal{M}}(q,\mathcal{P},p)$, where a smaller value indicates a
higher position.

Unlike free-form documents, each product $p$ is represented as a
structured listing consisting of fields such as title, features, and
description:
\[
p =
\left(
p^{\mathrm{title}},
p^{\mathrm{features}},
p^{\mathrm{description}}
\right).
\]
For a target product $p_t$, \textsc{Globe} generates a rewritten
listing $p'_t$ by applying editing behaviors whose permitted fields
are compatible with the product schema. Let $\mathcal{P}'$ denote the
candidate set in which $p_t$ is replaced by $p'_t$. We measure the
episode-level outcome using rank gain:
\[
g(q,p_t)
=
r_{\mathcal{M}}(q,\mathcal{P},p_t)
-
r_{\mathcal{M}}(q,\mathcal{P}',p'_t).
\]
The goal is to learn a handbook that produces positive rank gains across diverse query-product contexts while limiting harmful ranking outcomes.

\subsection{Behavior Handbook}
\label{sec:behavior}
A behavior is a reusable, field-scoped editing strategy for product listings.
Each behavior stores an \textbf{identifier, a name, an editing instruction, permitted target fields, conflicting behaviors, and feedback metadata}.
The \textit{instruction} defines a concrete editing operation, such as making the product type explicit or foregrounding a query-relevant specification.
The \textit{target-field} set specifies where the operation may be applied, with $\mathrm{targets}(b)\subseteq\{\mathrm{title},\mathrm{features},\mathrm{description}\}$.
The \textit{conflicts field} records behaviors that should not be applied together. 
Rather than strictly filtering these rules, this field serves as a soft constraint that instructs the editor to avoid applying conflicting behaviors simultaneously.
Finally, the \textit{feedback metadata} tracks the number of helpful, harmful, and neutral outcomes associated with behavior $b$, denoted by $h^{+}(b)$, $h^{-}(b)$, and $h^{0}(b)$, respectively.
We summarize the empirical utility of each behavior using the score $s(b)=h^{+}(b)-h^{-}(b)$.
Let $n(b)=h^{+}(b)+h^{-}(b)+h^{0}(b)$ denote the number of episodes in which the behavior has been applied.
To ensure that newly introduced behaviors receive sufficient opportunities for evaluation, a behavior $b$ remains active if it is still within its initial probation period ($n(b) < N_{\min}$) or maintains a non-negative empirical score ($s(b) \geq 0$). Behaviors with negative scores after the probation period remain stored for traceability but are excluded from the active behavior pool used in subsequent iterations. Examples of deactivated behaviors are listed in Appendix~\ref{app:deactivated}.

\subsection{Handbook Deployment}
At iteration $n$, the active behavior set $\mathcal{H}^{\mathrm{act}}_n$ is deployed across a batch of $K$ query-product pairs.
For each episode $(q_i,p_i)$, the LLM editor receives the user query, the original structured product listing, and the complete active behavior set as a strategy pool.
The editor selects the behaviors relevant to the current query-product context and generates a rewritten target product $p'_i$.
Each selected behavior may be applied only to the listing fields specified by its target-field scope.
The editor is also instructed to preserve the original section structure and leave fields outside the permitted scopes unchanged.
To record behavior usage, the editor also reports the subset of active behaviors applied during rewriting, denoted by $\mathcal{A}_i \subseteq \mathcal{H}^{\mathrm{act}}_n$.
Once the rewritten target product replaces the original product, the ranking model re-ranks the entire candidate set. 
Defining $r_i$ and $r'_i$ as the target product's rank before and after rewriting, respectively, the episode-level rank gain is calculated as:
\[
g_i=r_i-r'_i.
\]
A positive value indicates rank improvement, a negative value indicates degradation, and zero indicates no change.

The episode log therefore records the complete optimization trajectory, including the query, original and rewritten listings, reported behavior usage, and observed rank change.
Each episode is stored as
\[
e_i=\left(q_i,p_i,p'_i,\mathcal{A}_i,r_i,r'_i,g_i\right).
\]
At this stage, \textsc{Globe} only records the episode outcome and does not immediately update the handbook or behavior statistics.
All episode logs collected under $\mathcal{H}_n$ are aggregated after the $K$ episodes in the current batch have been completed.

\subsection{Handbook Evolution}
\label{sec:handbook_evolution}
After collecting all $K$ episodes under the current handbook $\mathcal{H}_n$, \textsc{Globe} aggregates their ranking outcomes and updates the handbook at the batch level.
For each episode, the $h^{+}$, $h^{-}$, or $h^{0}$ count of every applied behavior is incremented depending on whether the target product's rank improves, degrades, or remains unchanged, respectively.
Because the individual contribution of each behavior cannot be isolated, we assign the episode's overall directional outcome to all applied behaviors.
The accumulated statistics are then used to recompute each behavior's score and active status for the next iteration.
Using the aggregated episode logs and updated behavior statistics, \textsc{Globe} evolves the handbook through three components: 1) Reflector, 2) Moderator, and 3) Curator.
\paragraph{Reflector.}
The Reflector analyzes recent episode logs together with the current handbook to identify recurring success and failure patterns.
Rather than rewriting the handbook as unrestricted text, it proposes structured delta operations that either add a new behavior or modify an existing one.
An \textsc{Add} operation is proposed when the observed pattern is not adequately covered by the current handbook.
A \textsc{Modify} operation refines an existing behavior by clarifying its instruction, application condition, or permitted target fields.
The Reflector does not directly delete behaviors, since behaviors associated with consistently harmful outcomes are excluded through the active-status criterion while remaining stored for traceability.
\paragraph{Moderator.}
The Moderator filters Reflector proposals before they are committed to the handbook.
It performs \textbf{(1) embedding-based deduplication} to reject proposals that are semantically too similar to existing active behaviors.
A proposed behavior $b_{\mathrm{new}}$ is rejected when
\[
\max_{b\in\mathcal{H}^{act}_n}\operatorname{sim}\left(E(b_{\mathrm{new}}),E(b)\right)>\tau,
\]
where $E(\cdot)$ denotes the behavior embedding function and $\operatorname{sim}(\cdot,\cdot)$ denotes cosine similarity.
The remaining proposals undergo \textbf{(2) functional deduplication}, in which an LLM compares their application conditions and intended editing operations with those of existing behaviors.
When substantial overlap is detected, the existing behavior is retained, since unrestricted rewriting of committed behaviors would make the handbook fluctuate with the noise of individual batches.
\paragraph{Curator.}
The Curator deterministically commits the proposals accepted by the Moderator.
New behaviors receive stable identifiers and zero-initialized feedback metadata, while modified
behaviors are updated in place, advancing their version but retaining their accumulated metadata.
The Curator then constructs the next handbook $\mathcal{H}_{n+1}$ and determines its active behavior pool using the accumulated feedback statistics.
The updated handbook is applied to the next batch of query-product pairs, thereby repeating the rewriting, feedback collection, and handbook evolution cycle.
\subsection{Training and Inference}
\label{sec:inference}
Starting from a seed handbook $\mathcal{H}_1$ containing nine general GEO heuristics, \textsc{Globe} repeats the\textit{ deployment-and-evolution} cycle for $N$ training batches. 
The handbook and its active behavior pool remain fixed within each batch. 
Feedback collected during the batch affects behavior status and handbook contents only in the subsequent iteration, ensuring that all episodes in the same batch are generated under a consistent handbook version.
After the final training batch, the handbook is frozen. At test time, the active behaviors from the frozen handbook are supplied to the editor using the same deployment procedure as during training. 
The editor selects and applies behaviors for each query-product pair, but no feedback statistics or handbook contents are updated during evaluation.

\section{Experiment}
\begin{table*}[t]
\centering
\small
\setlength{\tabcolsep}{5pt}
\begin{tabular}{lccc cc cc c}
\toprule
\multirow{2}{*}{\textbf{Method}}
& \multicolumn{3}{c}{\textbf{Components}}
& \multicolumn{2}{c}{\textbf{Effectiveness}}
& \multicolumn{2}{c}{\textbf{Risk \& Consistency}}
& \multirow{2}{*}{\textbf{Handbook Size}} \\
\cmidrule(lr){2-4}
\cmidrule(lr){5-6}
\cmidrule(lr){7-8}
& Handbook
& Feedback
& Moderator
& MRG$\uparrow$
& SR$\uparrow$
& DSR$\downarrow$
& CV$\downarrow$
& \\
\midrule
Prompt Opt.
& --
& --
& --
& 0.351
& 44.1\%
& 3.961
& 0.117
& -- \\
Static HB
& \checkmark
& --
& --
& 1.343
& 58.6\%
& 2.628
& \underline{0.042}
& 9 \\
\textsc{Globe} w/o Mod.
& \checkmark
& \checkmark
& --
& \underline{1.519}
& \underline{60.6\%}
& \underline{2.389}
& 0.054
& 51 \\
\rowcolor{gray!10}
\textsc{\textbf{\textsc{Globe}}}
& \checkmark
& \checkmark
& \checkmark
& \textbf{1.611}
& \textbf{61.6\%}
& \textbf{2.172}
& \textbf{0.034}
& 47 \\
\bottomrule
\end{tabular}
\caption{Performance comparison of optimization methods. Higher values are better for MRG and SR ($\uparrow$), while lower values are better for DSR and CV ($\downarrow$).
Best results are shown in bold and second-best results are underlined.}
\label{tab:main_results}
\end{table*}

\subsection{Experimental Setup}
\label{subsec:experimental_setup}

\paragraph{Dataset.} 
We evaluate \textsc{Globe} using the E-GEO benchmark \citep{bagga2025egeo}, which pairs real-world shopping queries from Reddit (r/BuyItForLife) with structured Amazon product listings.
We categorize the products into six macro-categories: Home \& Kitchen, Fashion \& Accessories, Electronics, Tools \& DIY, Sports \& Outdoors, and Health \& Beauty.
We use 1,000 training queries to construct the behavior handbook and a disjoint, category-balanced set of 900 queries (150 per category) for evaluation.
The handbook remains frozen during testing.

\paragraph{Baselines.}
We evaluate \textsc{Globe} against the following configurations.\footnote{Recent self-evolving agents are not directly applicable as our baselines because they are fundamentally designed for free-form document optimization, rendering field-scoped constraints incompatible with their core exploration mechanism.}
\textbf{Prompt Opt.} applies a single monolithic rewriting instruction following the prompt-based E-GEO setting~\citep{bagga2025egeo}.
\textbf{Static HB} utilizes a fixed, non-evolving handbook of 9 generic rewrite rules~\citep{aggarwal2023geo}.
\textbf{GLOBE w/o Mod.} evolves the handbook without Moderator, using episode feedback but commits Reflector proposals without redundancy filtering.

\paragraph{Evaluation Metrics.}
We evaluate system performance using four metrics.
\textbf{Mean Rank Gain (MRG)} and \textbf{Success Rate (SR)} measure the average position improvement and the frequency of successful optimizations, respectively.
To evaluate stability, we evaluate \textbf{Downside Risk (DSR)}, which applies a quadratic penalty to rank degradations to capture the severity of harmful rewrites.
Finally, \textbf{Category Variance (CV)} measures the variance of MRG across the six categories, where a lower score indicates generalized performance.
More details are presented in Appendix~\ref{app:metrics}.

\paragraph{Training Details.}
The initial handbook contains 9 general GEO heuristics in Appendix~\ref{app:seed}.
We process the 1,000 training queries in 10 batches of 100 query-product pairs.
Within each batch, the current handbook remains fixed while product rewrites and episode logs are collected.
After all episodes in the batch have been completed, behavior statistics are accumulated and the handbook is updated through the Reflector, Moderator, and Curator.
We use GPT-5.4\footnote{https://platform.openai.com/docs/models/gpt-5.4} for all LLM-based components.
Embedding-based deduplication uses \texttt{all-MiniLM-L6-v2}~\citep{reimers-gurevych-2019-sentence,3495724.3496209} with a cosine-similarity threshold of $\tau=0.9$.
More details and prompts are provided in Appendix~\ref{app:details} and~\ref{app:prompts}.

\subsection{Effectiveness of the Behavior Handbook}
\label{subsec:effectiveness}

To evaluate the effectiveness of our proposed handbook framework, we examine the contribution of its core design components to the overall performance of \textsc{Globe}.
As shown in Table~\ref{tab:main_results}, the Static HB already achieves a substantial improvement over monolithic prompt optimization in both ranking effectiveness and downside risk.
Because this baseline uses only the initial generic behaviors without feedback-based updates, its improvement suggests that decomposing optimization knowledge into explicit operations, together with field scoping, provides a stronger foundation than the evaluated monolithic prompt baseline. This result further shows that structured behavior representation contributes substantially to \textsc{Globe}’s performance even before feedback-driven evolution is introduced.

Feedback-based evolution further improves MRG and SR by extending the handbook beyond its initial generic heuristics.
However, the unmoderated variant also produces a larger handbook and higher category variance, suggesting that unrestricted accumulation may introduce redundant or narrowly specialized behaviors.
The increase in handbook capacity also means that its improvement cannot be attributed entirely to feedback-based learning.
\textbf{Evolution expands the available optimization knowledge, but accumulating more behaviors does not necessarily produce a more consistent handbook.}

The full \textsc{Globe} addresses this limitation by filtering semantically redundant and functionally overlapping proposals before they are committed.
Despite retaining fewer behaviors than the unmoderated variant, it achieves stronger overall performance together with lower DSR and CV.
This pattern suggests that the Moderator acts as handbook-level regularization, improving the composition of the accumulated knowledge rather than simply expanding its size.
\textbf{Overall, moderated evolution produces a more compact handbook with a better balance between ranking effectiveness and consistency.}

\subsection{Impact of Feedback-driven Refinement}
\label{subsec:refinement}
\begin{figure}[t]
\centering
\includegraphics[width=\columnwidth]{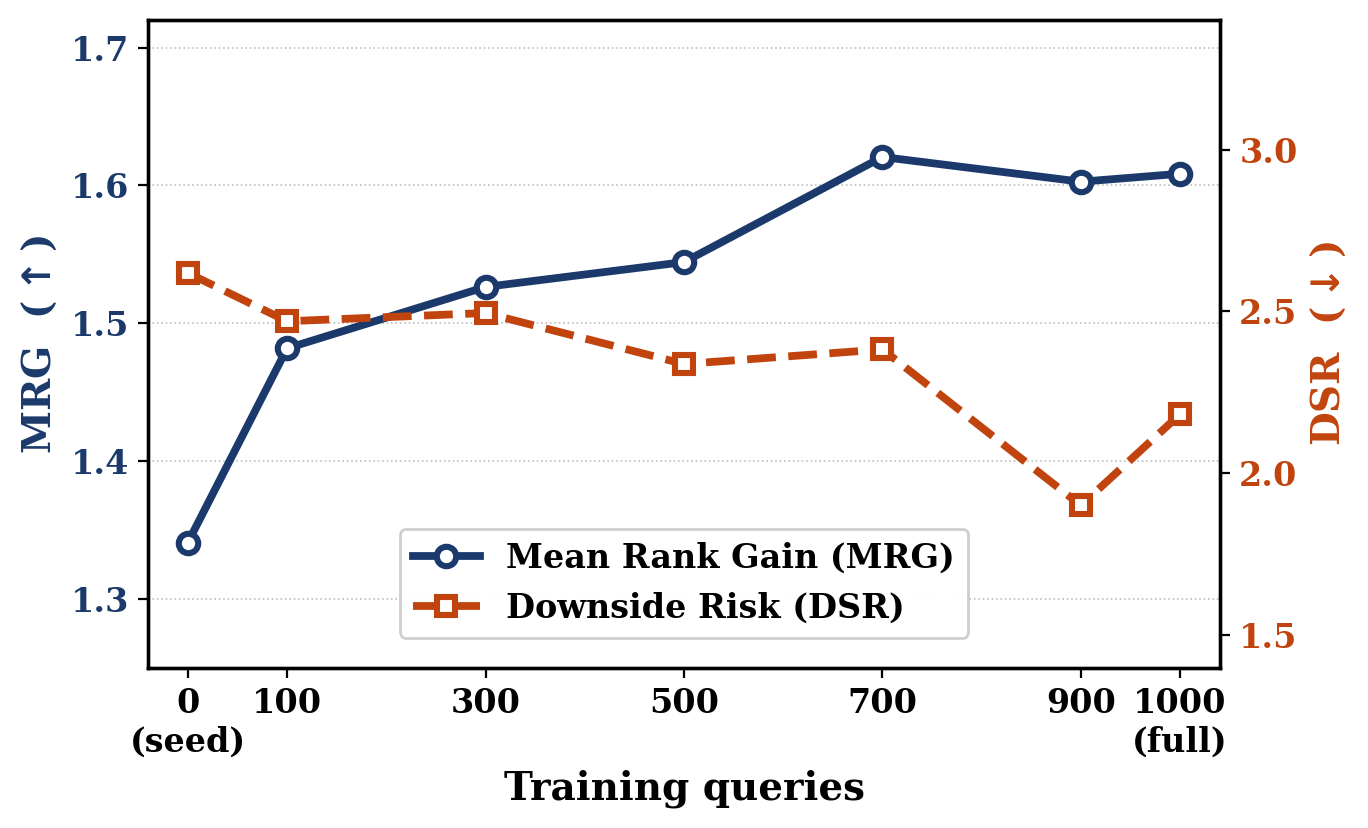}
\caption{Performance of the trained handbook as the number of training queries increases. }
\label{fig:performance_trajectory}
\end{figure}
To examine whether iterative handbook evolution produces sustained improvements beyond the final aggregate results, we trace ranking performance as the number of training examples increases.

As shown in Figure~\ref{fig:performance_trajectory}, MRG improves rapidly in the early stages of training, as feedback from the initial batches reveals optimization patterns not covered by the seed handbook, and then plateaus as additional batches are processed.
This trajectory indicates that \textbf{feedback-based evolution progressively enriches the handbook, but its benefits exhibit diminishing returns once broadly useful behaviors have been acquired.}
DSR also declines overall, though not monotonically, indicating that evolution
gradually mitigates harmful ranking outcomes.
Together, these patterns suggest that once frequently occurring optimization
patterns have been observed, further improvement depends less on training-set
size than on \textbf{broader coverage of query intents and product contexts
that provide genuinely new feedback}.

Additionally, to test whether refinement improves the specific behaviors it targets,
rather than merely coinciding with handbook growth, we compare each
modified behavior's episode-level outcomes before and after modification
(Appendix~\ref{app:modify_prepost}). Among the nine under-performing behaviors that were modified, eight showed improvements beyond a matched mean-reversion control (sign test, p = 0.020), with an average adjusted increase of 5.3 percentage points in helpful rate. 
In contrast, modifying behaviors that were already performing above average yields no measurable gain.
These results suggest that \textbf{\textsc{modify} operations act as targeted repairs for under-performing behaviors rather than as indiscriminate handbook expansion.}

\subsection{Cross-Category Generalizability}
\begin{table*}[t]
\centering
\resizebox{\textwidth}{!}{
\begin{tabular}{lcccccccccccc}
\toprule
\multirow{2}{*}{\textbf{Method}} & \multicolumn{4}{c}{\textbf{GPT-5.4 (Self)}} & \multicolumn{4}{c}{\textbf{Qwen3.5-122B-A10B}} & \multicolumn{4}{c}{\textbf{Gemini-3.5-flash}} \\
\cmidrule(lr){2-5} \cmidrule(lr){6-9} \cmidrule(lr){10-13}
 & \textbf{MRG$\uparrow$} & \textbf{SR$\uparrow$} & \textbf{DSR$\downarrow$} & \textbf{CV$\downarrow$} & \textbf{MRG$\uparrow$} & \textbf{SR$\uparrow$} & \textbf{DSR$\downarrow$} & \textbf{CV$\downarrow$} & \textbf{MRG$\uparrow$} & \textbf{SR$\uparrow$} & \textbf{DSR$\downarrow$} & \textbf{CV$\downarrow$} \\
\midrule
Prompt Opt. & 0.351 & 44.1\% & 3.961 & 0.117 & $-$0.031 & 22.1\% & 1.700 & 0.044 & $-$0.051 & 21.5\% & 1.379 & 0.019 \\
Static HB & 1.343 & 58.6\% & 2.628 & 0.042 & 1.335 & 50.3\% & 0.435 & 0.048 & 0.960 & 45.5\% & 0.290 & 0.015 \\
\textbf{\textsc{Globe}} w/o Mod. & 1.519 & 60.6\% & 2.389 & 0.054 & 1.167 & 46.6\% & 0.115 & \textbf{0.010} & 0.975 & 46.6\% & 0.371 & 0.021 \\
\rowcolor{gray!10}
\textbf{\textsc{Globe}} & \textbf{1.611} & \textbf{61.6\%} & \textbf{2.172} & \textbf{0.034} & \textbf{1.522} & \textbf{54.1\%} & \textbf{0.105} & 0.034 & \textbf{1.078} & \textbf{48.9\%} & \textbf{0.246} & \textbf{0.013} \\
\bottomrule
\end{tabular}
}
\caption{Cross-ranker evaluation of optimization strategies trained with GPT-5.4.}
\label{tab:cross_ranker}
\end{table*}
\begin{figure}[t]
\centering
\includegraphics[width=\columnwidth]{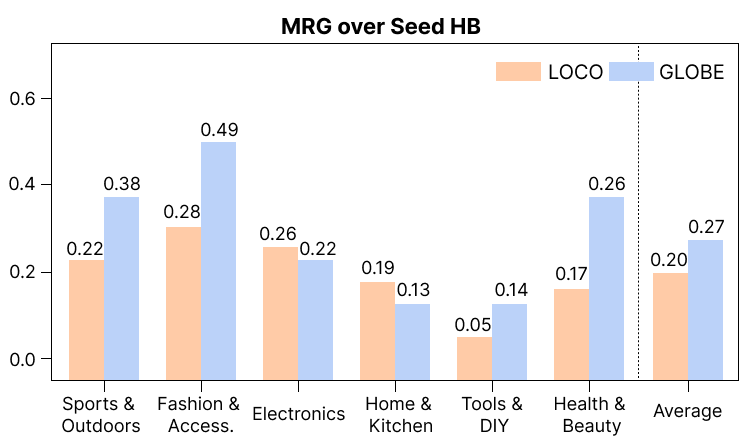}
\caption{Cross-category generalization, measured as MRG over the seed handbook.}
\label{fig:loco_generalization}
\end{figure}
\label{subsec:generalizability}

To examine whether the evolved handbook captures transferable optimization knowledge rather than category-specific heuristics, we conduct a leave-one-category-out (LOCO) evaluation.
Specifically, for each macro-category, we evolve a separate handbook using only the training queries from the remaining five categories.
Figure~\ref{fig:loco_generalization} compares the MRG improvement over the seed handbook for the LOCO models and the fully trained \textsc{Globe} model.
The result confirms that the benefits of feedback-driven evolution are not strictly bound to domain-specific exposure.
The LOCO handbook consistently retains a substantial portion of the aggregate improvements seen in the fully trained model.
Furthermore, the LOCO approach even surpasses full training in some categories.
These category-level differences suggest that training on all categories can occasionally hurt a specific category through cross-category interference, which we leave for future analysis.
Nevertheless, they show that in-category exposure is not a strict prerequisite for generating effective rewrites.
Together with the modest overall MRG decrease of 0.07, these results suggest that \textbf{feedback-driven evolution extracts transferable optimization
strategies that remain effective without target-category exposure, avoiding category-specific memorization.}

\subsection{Cross-Ranker Transferability}
\label{sec:cross_ranker}
\newcommand{\cshead}[1]{\addlinespace[1pt]%
  \multicolumn{2}{@{}l@{}}{\textit{#1}} \\ \addlinespace[2pt]}
\newcommand{\btag}[1]{\,{\footnotesize[#1]}}

\begin{table}[t]
\centering
\small
\setlength{\tabcolsep}{3pt}
\renewcommand{\arraystretch}{1.08}
\begin{tabular}{@{}>{\raggedright\arraybackslash}p{1.3cm}>{\raggedright\arraybackslash}p{5.9cm}@{}}
\toprule
\multicolumn{2}{@{}p{7.3cm}@{}}{\textbf{Query:} \textit{``Aux cable for the car that won't break after a month of use.''} \;\textbf{Target Field:} \textit{features}} \\
\midrule
\cshead{Delta operations committed into $\mathcal{H}_4$}
\textsc{modify} \newline B012 &
Name the listing's material and build detail \textbf{most relevant to the likely wear point}
\newline \textit{(from:} name the listing's material and build \textit{details)} \\
\addlinespace[4pt]
\textsc{add} \newline B016 &
When the query \textbf{describes how past products failed}, add a top-placed clause mapping
construction evidence to \textbf{that exact failure mode} \\
\midrule
\cshead{Resulting feature bullets}
$\mathcal{H}_2$ &
\textit{rank $4 \rightarrow 5$, $g_i = -1$}
\newline (i) tight spaces and \textbf{connector strain} often cause early cable failure
\newline (ii) reduces \textbf{bend stress at the jack} for better day-to-day durability \\
\addlinespace[4pt]
$\mathcal{H}_4$ &
\textit{rank $4 \rightarrow 0$, $g_i = +4$}
\newline (i) tight spaces and \textbf{repeated bending} often cause cable failure\btag{B016}
\newline (ii) reduces strain \textbf{at the connector to address a common breakage
point}\btag{B012} \\
\midrule
\multicolumn{2}{@{}p{7.2cm}@{}}{\textbf{Downstream:} B016 is applied in 185 subsequent episodes
(62\% helpful, 13\% harmful, mean gain $+1.90$).} \\
\bottomrule
\end{tabular}
\caption{Behavior evolution from feedback to ranking effect for the target product \textit{CableCreation 6\,FT Flat 3.5mm Aux Cable}. Full example in Appendix~\ref{sec:case_study_app}.}
\label{tab:case-compact}
\end{table}

To determine whether the observed rank gains stem from transferable optimization rather than a generative engine simply favoring its own outputs, we freeze the GPT-5.4 editor and the handbook it produced.
We then replace the ranking model with two different engines: Qwen3.5-122B-A10B\footnote{https://huggingface.co/Qwen/Qwen3.5-122B-A10B} and Gemini-3.5-flash\footnote{https://ai.google.dev/gemini-api/docs/models/gemini-3.5-flash}.
Although all rankers operate on the same ten-position rank scale, they differ in baseline orderings, calibration, and ranking stability. 
We therefore emphasize within-ranker comparisons between methods rather than interpreting absolute differences across rankers as model-quality differences.
As shown in Table~\ref{tab:cross_ranker}, the best and worst methods are unchanged across all three rankers.
\textsc{Globe} achieves the highest MRG and SR, whereas Prompt Opt.\ remains the weakest baseline, with its gains turning negative when evaluated by the held-out rankers.
This suggests that prompt optimization may overfit to the specific preferences of its training engine.
The two handbook variants, however, diverge.
\textsc{Globe} consistently outperforms the static seed handbook across all rankers, whereas the unmoderated variant loses its clear advantage over the static baseline on unseen rankers.
Consequently, the performance gap between \textsc{Globe} and \textsc{Globe} w/o Mod.\ widens on the unseen rankers, where \textsc{Globe} also demonstrates the lowest downside risk.
These findings suggest that moderation provides benefits beyond mere handbook compactness.
Without filtering, feedback-driven accumulation absorbs ranker-specific biases that fail to transfer, whereas \textbf{moderated evolution retains behaviors that remain effective across different judges.}

\subsection{Case Study}
\label{sec:case_study}
Table~\ref{tab:case-compact} shows a single handbook evolution from ranking feedback to its downstream editing effect. Under $\mathcal{H}_2$, the existing durability behavior (B012) degraded the target product's rank ($g_i = -1$) because it merely surfaced generic material evidence without addressing the shopper's specific concern. In response, the Reflector proposes a \textsc{modify} to narrow B012 to wear-point specific details, along with an \textsc{add} for a new behavior (B016) that explicitly maps construction evidence to the user's stated failure mode. Once the Curator commits these deltas into $\mathcal{H}_4$, the editor applying the updated handbook ($\mathcal{H}_4$) stops merely listing specs; instead, it directly targets \textit{"repeated bending"} and \textit{"connector"} strain. This strategic shift improves the product from rank 4 to the top position ($g_i = +4$). Furthermore, B016 generalizes well beyond this initial episode, achieving a 62\% helpful rate and a $+1.90$ mean rank gain across 185 subsequent applications. \textbf{\textsc{Globe} therefore renders optimization knowledge traceable by explicitly linking observed failures to structural behavior updates and their downstream ranking improvements.}

\section{Conclusion}
We introduced \textsc{Globe}, a framework that maintains product-optimization knowledge as explicit, field-scoped behaviors in an evolving handbook. By converting ranking feedback into persistent and reusable editing knowledge, \textsc{Globe} enables effective and interpretable optimization while transferring to unseen product categories. More broadly, our findings suggest that evolving structured behaviors offers a practical path toward interpretable and transferable agentic memory for generative search.

\section*{Limitations}
A key limitation is that the editor, ranker, Reflector, and Moderator are all instantiated from the same model (GPT-5.4). Because the ranker that provides the reward signal shares a model family with the components optimized against it, the learned behaviors may partly capture ranker-specific preferences rather than ranker-agnostic optimization knowledge. We probe this with a cross-ranker evaluation in Section~\ref{sec:cross_ranker}; fully disentangling transferable knowledge from ranker-specific artifacts remains future work.
In addition, credit assignment over behaviors is approximate. Since each episode yields a single rank change for the entire rewrite, the episode's directional outcome is attributed to all applied behaviors rather than to each behavior's individual contribution (Section~\ref{sec:handbook_evolution}). 
A useful behavior can thus absorb harmful counts from co-applied behaviors.
These counts also rely on the editor's self-reported list of applied behaviors, which is not verified against the actual edits; an audit of 50 training episodes judged by a separate LLM (Claude Opus 5) found 88.3\% of reported behaviors consistent with the edits made and few applied behaviors left unreported.
Batch-level aggregation and the probation period mitigate this attribution noise but do not eliminate it. More precise attribution, such as per-behavior counterfactual rewrites, would incur substantially higher ranking costs per episode, and we leave its efficient integration into the evolution loop to future work.
Moreover, because modified behaviors inherit their accumulated feedback counts, the active-status criterion reacts slowly to a harmful \textsc{modify}, and \textsc{Globe} currently has no mechanism to revert such a revision.
Beyond these signal-level limitations, while establishing a rigorous performance ceiling with high-end models necessitates substantial computational overhead during the multi-agent learning phase, our inference stage remains highly efficient; the compiled handbook acts as a static set of rules that applies verified behaviors in a single pass without iterative optimization loops. Furthermore, distilling these structured behaviors into smaller, open-source models via fine-tuning represents a promising direction to drastically reduce future deployment costs.
Finally, as real-world e-commerce search algorithms are dynamic and continuously updated, accumulated editing behaviors may experience temporal degradation and require periodic re-evolution.
Extending our current text-only, field-scoped constraints to multimodal settings also remains an open avenue for future research.

\section*{Ethical Considerations}
\paragraph{Factuality and Consumer Protection.}
Methods for improving product visibility in generative
search may be used to unfairly influence rankings, and
rank improvement should not be interpreted as evidence
of superior product quality. To reduce misleading edits,
\textsc{Globe} includes a factual-grounding behavior that
requires rewritten claims to remain supported by the
original listing. Nevertheless, language models may still
introduce unsupported implications or overstate product
benefits. In practical deployments, rewritten listings
should therefore be verified against the original listing
and monitored for deceptive or exaggerated claims.

\paragraph{Model Bias and Accessibility.}
The ranking feedback used by \textsc{Globe} may reflect
model-specific preferences rather than general shopper
preferences. We partially examine this risk through
cross-ranker evaluation, but it cannot be fully eliminated.
In addition, evolving the handbook with high-capacity
language models may favor organizations with greater
computational resources. Reusing a frozen handbook and
adapting it to smaller models may help reduce this barrier.

\bibliography{custom}

@article{aggarwal2023geo,
  title   = {{GEO}: Generative Engine Optimization},
  author  = {Aggarwal, Pranjal and Murahari, Vishvak and Rajpurohit, Tanmay and Kalyan, Ashwin and Narasimhan, Karthik and Deshpande, Ameet},
  journal = {arXiv preprint arXiv:2311.09735},
  year    = {2023}
}

@article{kumar2024manipulating,
  title   = {Manipulating Large Language Models to Increase Product Visibility},
  author  = {Kumar, Aounon and Lakkaraju, Himabindu},
  journal = {arXiv preprint arXiv:2404.07981},
  year    = {2024}
}

@article{wu2025autogeo,
  title   = {What Generative Search Engines Like and How to Optimize Web Content Cooperatively},
  author  = {Wu, Yujiang and Zhong, Shanshan and Kim, Yubin and Xiong, Chenyan},
  journal = {arXiv preprint arXiv:2510.11438},
  year    = {2025}
}

@article{bagga2025egeo,
  title   = {{E-GEO}: A Testbed for Generative Engine Optimization in E-Commerce},
  author  = {Bagga, Puneet S. and Farias, Vivek F. and Korkotashvili, Tamar and Peng, Tianyi and Wu, Yuhang},
  journal = {arXiv preprint arXiv:2511.20867},
  year    = {2025}
}

@article{jin2026core,
  title   = {Controlling Output Rankings in Generative Engines for {LLM}-based Search},
  author  = {Jin, Haibo and Chen, Ruoxi and Zhang, Peiyan and Luo, Yifeng and Zeng, Huimin and Luo, Man and Wang, Haohan},
  journal = {arXiv preprint arXiv:2602.03608},
  year    = {2026}
}

@article{yuan2026agenticgeo,
  title   = {{AgenticGEO}: A Self-Evolving Agentic System for Generative Engine Optimization},
  author  = {Yuan, Jiaqi and Wang, Jialu and Wang, Zihan and Sun, Qingyun and Wang, Ruijie and Li, Jianxin},
  journal = {arXiv preprint arXiv:2603.20213},
  year    = {2026}
}

@article{madaan2023self,
  title={Self-refine: Iterative refinement with self-feedback},
  author={Madaan, Aman and Tandon, Niket and Gupta, Prakhar and Hallinan, Skyler and Gao, Luyu and Wiegreffe, Sarah and Alon, Uri and Dziri, Nouha and Prabhumoye, Shrimai and Yang, Yiming and others},
  journal={Advances in neural information processing systems},
  volume={36},
  pages={46534--46594},
  year={2023}
}

@article{schick2022peer,
  title={Peer: A collaborative language model},
  author={Schick, Timo and Dwivedi-Yu, Jane and Jiang, Zhengbao and Petroni, Fabio and Lewis, Patrick and Izacard, Gautier and You, Qingfei and Nalmpantis, Christoforos and Grave, Edouard and Riedel, Sebastian},
  journal={arXiv preprint arXiv:2208.11663},
  year={2022}
}

@article{shinn2023reflexion,
  title={Reflexion: Language agents with verbal reinforcement learning},
  author={Shinn, Noah and Cassano, Federico and Gopinath, Ashwin and Narasimhan, Karthik and Yao, Shunyu},
  journal={Advances in neural information processing systems},
  volume={36},
  pages={8634--8652},
  year={2023}
}

@inproceedings{zhao2024expel,
  title={Expel: Llm agents are experiential learners},
  author={Zhao, Andrew and Huang, Daniel and Xu, Quentin and Lin, Matthieu and Liu, Yong-Jin and Huang, Gao},
  booktitle={Proceedings of the AAAI Conference on Artificial Intelligence},
  volume={38},
  number={17},
  pages={19632--19642},
  year={2024}
}

@article{didolkar2025metacognitive,
  title={Metacognitive reuse: Turning recurring llm reasoning into concise behaviors},
  author={Didolkar, Aniket and Ballas, Nicolas and Arora, Sanjeev and Goyal, Anirudh},
  journal={arXiv preprint arXiv:2509.13237},
  year={2025}
}

@article{zhang2025agentic,
  title={Agentic context engineering: Evolving contexts for self-improving language models},
  author={Zhang, Qizheng and Hu, Changran and Upasani, Shubhangi and Ma, Boyuan and Hong, Fenglu and Kamanuru, Vamsidhar and Rainton, Jay and Wu, Chen and Ji, Mengmeng and Li, Hanchen and others},
  journal={arXiv preprint arXiv:2510.04618},
  year={2025}
}

@article{xiong2026learning,
  title={Learning to continually learn via meta-learning agentic memory designs},
  author={Xiong, Yiming and Hu, Shengran and Clune, Jeff},
  journal={arXiv preprint arXiv:2602.07755},
  year={2026}
}

@article{zhou2026if,
  title={IF-GEO: Conflict-Aware Instruction Fusion for Multi-Query Generative Engine Optimization},
  author={Zhou, Heyang and Chen, JiaJia and Chen, Xiaolu and Bao, Jie and Chen, Zhen and Liao, Yong},
  journal={arXiv preprint arXiv:2601.13938},
  year={2026}
}

@inproceedings{li-etal-2024-large,
    title = "Large Language Models for Generative Recommendation: A Survey and Visionary Discussions",
    author = "Li, Lei  and
      Zhang, Yongfeng  and
      Liu, Dugang  and
      Chen, Li",
    editor = "Calzolari, Nicoletta  and
      Kan, Min-Yen  and
      Hoste, Veronique  and
      Lenci, Alessandro  and
      Sakti, Sakriani  and
      Xue, Nianwen",
    booktitle = "Proceedings of the 2024 Joint International Conference on Computational Linguistics, Language Resources and Evaluation (LREC-COLING 2024)",
    month = may,
    year = "2024",
    address = "Torino, Italia",
    publisher = "ELRA and ICCL",
    url = "https://aclanthology.org/2024.lrec-main.886/",
    pages = "10146--10159"
}

@inproceedings{sun-etal-2023-chatgpt,
    title = "Is {C}hat{GPT} Good at Search? Investigating Large Language Models as Re-Ranking Agents",
    author = "Sun, Weiwei  and
      Yan, Lingyong  and
      Ma, Xinyu  and
      Wang, Shuaiqiang  and
      Ren, Pengjie  and
      Chen, Zhumin  and
      Yin, Dawei  and
      Ren, Zhaochun",
    editor = "Bouamor, Houda  and
      Pino, Juan  and
      Bali, Kalika",
    booktitle = "Proceedings of the 2023 Conference on Empirical Methods in Natural Language Processing",
    month = dec,
    year = "2023",
    address = "Singapore",
    publisher = "Association for Computational Linguistics",
    url = "https://aclanthology.org/2023.emnlp-main.923/",
    doi = "10.18653/v1/2023.emnlp-main.923",
    pages = "14918--14937"
}

@inproceedings{hou2024large,
  title={Large language models are zero-shot rankers for recommender systems},
  author={Hou, Yupeng and Zhang, Junjie and Lin, Zihan and Lu, Hongyu and Xie, Ruobing and McAuley, Julian and Zhao, Wayne Xin},
  booktitle={European conference on information retrieval},
  pages={364--381},
  year={2024},
  organization={Springer}
}

@inproceedings{reimers-gurevych-2019-sentence,
    title = "Sentence-{BERT}: Sentence Embeddings using {S}iamese {BERT}-Networks",
    author = "Reimers, Nils  and
      Gurevych, Iryna",
    editor = "Inui, Kentaro  and
      Jiang, Jing  and
      Ng, Vincent  and
      Wan, Xiaojun",
    booktitle = "Proceedings of the 2019 Conference on Empirical Methods in Natural Language Processing and the 9th International Joint Conference on Natural Language Processing (EMNLP-IJCNLP)",
    month = nov,
    year = "2019",
    address = "Hong Kong, China",
    publisher = "Association for Computational Linguistics",
    url = "https://aclanthology.org/D19-1410/",
    doi = "10.18653/v1/D19-1410",
    pages = "3982--3992"
}

@misc{agrawal2026gepareflectivepromptevolution,
      title={GEPA: Reflective Prompt Evolution Can Outperform Reinforcement Learning}, 
      author={Lakshya A Agrawal and Shangyin Tan and Dilara Soylu and Noah Ziems and Rishi Khare and Krista Opsahl-Ong and Arnav Singhvi and Herumb Shandilya and Michael J Ryan and Meng Jiang and Christopher Potts and Koushik Sen and Alexandros G. Dimakis and Ion Stoica and Dan Klein and Matei Zaharia and Omar Khattab},
      year={2026},
      eprint={2507.19457},
      archivePrefix={arXiv},
      primaryClass={cs.CL},
      url={https://arxiv.org/abs/2507.19457}, 
}

@inproceedings{pryzant-etal-2023-automatic,
    title = "Automatic Prompt Optimization with ``Gradient Descent'' and Beam Search",
    author = "Pryzant, Reid  and
      Iter, Dan  and
      Li, Jerry  and
      Lee, Yin  and
      Zhu, Chenguang  and
      Zeng, Michael",
    editor = "Bouamor, Houda  and
      Pino, Juan  and
      Bali, Kalika",
    booktitle = "Proceedings of the 2023 Conference on Empirical Methods in Natural Language Processing",
    month = dec,
    year = "2023",
    address = "Singapore",
    publisher = "Association for Computational Linguistics",
    url = "https://aclanthology.org/2023.emnlp-main.494/",
    doi = "10.18653/v1/2023.emnlp-main.494",
    pages = "7957--7968"
}

@article{yuksekgonul2024textgrad,
  title={Textgrad: Automatic" differentiation" via text},
  author={Yuksekgonul, Mert and Bianchi, Federico and Boen, Joseph and Liu, Sheng and Huang, Zhi and Guestrin, Carlos and Zou, James},
  journal={arXiv preprint arXiv:2406.07496},
  year={2024}
}

@inproceedings{3495724.3496209,
author = {Wang, Wenhui and Wei, Furu and Dong, Li and Bao, Hangbo and Yang, Nan and Zhou, Ming},
title = {MINILM: deep self-attention distillation for task-agnostic compression of pre-trained transformers},
year = {2020},
isbn = {9781713829546},
publisher = {Curran Associates Inc.},
address = {Red Hook, NY, USA},
booktitle = {Proceedings of the 34th International Conference on Neural Information Processing Systems},
articleno = {485},
numpages = {13},
location = {Vancouver, BC, Canada},
series = {NIPS '20}
}

@book{davis2006search,
  title={Search engine optimization},
  author={Davis, Harold},
  year={2006},
  publisher={" O'Reilly Media, Inc."}
}

@article{10.1109/MC.2015.297,
author = {Gudivada, Venkat N. and Rao, Dhana and Paris, Jordan},
title = {Understanding Search-Engine Optimization},
year = {2015},
issue_date = {Oct. 2015},
publisher = {IEEE Computer Society Press},
address = {Washington, DC, USA},
volume = {48},
number = {10},
issn = {0018-9162},
url = {https://doi.org/10.1109/MC.2015.297},
doi = {10.1109/MC.2015.297},
journal = {Computer},
month = oct,
pages = {43–52},
numpages = {10}
}

@article{wang2023voyager,
  title={Voyager: An open-ended embodied agent with large language models},
  author={Wang, Guanzhi and Xie, Yuqi and Jiang, Yunfan and Mandlekar, Ajay and Xiao, Chaowei and Zhu, Yuke and Fan, Linxi and Anandkumar, Anima},
  journal={arXiv preprint arXiv:2305.16291},
  year={2023}
}

@inproceedings{10.1145/3726302.3730055,
author = {Kim, Jieyong and Kim, Hyunseo and Cho, Hyunjin and Kang, SeongKu and Chang, Buru and Yeo, Jinyoung and Lee, Dongha},
title = {Review-driven Personalized Preference Reasoning with Large Language Models for Recommendation},
year = {2025},
isbn = {9798400715921},
publisher = {Association for Computing Machinery},
address = {New York, NY, USA},
url = {https://doi.org/10.1145/3726302.3730055},
doi = {10.1145/3726302.3730055},
booktitle = {Proceedings of the 48th International ACM SIGIR Conference on Research and Development in Information Retrieval},
pages = {1697–1706},
numpages = {10},
location = {Padua, Italy},
series = {SIGIR '25}
}

@article{kim2025bespoke,
 title={BESPOKE: Benchmark for search-augmented large language model personalization via diagnostic feedback},
 author={Kim, Hyunseo and Lee, Sangam and Seo, Kwangwook and Lee, Dongha},
 journal={arXiv preprint arXiv:2509.21106},
 year={2025}
}

\appendix
\label{sec:appendix}
\section{Robustness Across Different LLMs}
\label{sec:llm_backends}
\begin{table}[ht]
\centering
\small
\begin{tabularx}{\columnwidth}{l *{5}{>{\centering\arraybackslash}X}}
\toprule
\textbf{Model} & \textbf{MRG$\uparrow$} & \textbf{SR$\uparrow$} & \textbf{DSR$\downarrow$} & \textbf{CV$\downarrow$} & \textbf{HB} \\
\midrule
GPT-5.4 & 1.611 & \textbf{61.6\%} & \textbf{2.172} & 0.034 & 47 \\
Qwen3.5-122B-A10B & 1.507 & 56.5\% & 2.106 & 0.058 & 29 \\
\textbf{Gemini-3.5-flash} & \textbf{1.700} & 60.8\% & 2.310 & \textbf{0.028} & 46 \\
\bottomrule
\end{tabularx}
\caption{Performance evaluation of the GLOBE framework across different LLM backends.}
\label{tab:llm_backends}
\end{table}
To assess backbone robustness, we evaluate \textsc{Globe} across multiple LLMs.
For each backbone, we rerun the complete handbook-evolution procedure from the seed handbook, using that model for all LLM-based components and retaining the same data splits and hyperparameters.
As shown in Table~\ref{tab:llm_backends}, all configurations achieve positive rank gains, while exhibiting different trade-offs across MRG, SR, and DSR.
These results suggest that \textsc{Globe} remains effective across both proprietary and open-weight models, although its performance depends on the underlying backbone.

\section{Experimental Details}
\label{app:details}

\subsection{Dataset and Evaluation Setup}
\label{app:dataset}

Every method optimizes the same target product in the same competitive context, and is scored
against the same reference ranking. Each E-GEO query~\citep{bagga2025egeo} comes with a candidate
set of ten Amazon listings, each serialized into the title, features, and description fields. We
sample one candidate uniformly at random as the target product, store that index once, and reuse it
for GLOBE, its ablations, and all baselines; the pre-rewrite ranking
$r_{\mathcal{M}}(q,\mathcal{P},p_t)$ is likewise computed once and cached. 
This shared setup is intended to isolate differences arising from the rewriting strategy.

We use 1{,}000 queries for handbook construction, processed as ten batches of 100 pairs, and a
disjoint set of 900 queries for evaluation. 
Category labels are not provided by the benchmark at the query level. 
We first use GPT-5.4-nano to construct a mapping table from the unique Amazon main category values in the dataset to the six macro-categories. 
This mapping table is applied deterministically to every candidate product. 
Each query is assigned the majority macro-category among the candidates in its product set.
The evaluation set is built by sampling 150 queries per macro-category.
An episode is discarded when the editor returns no usable rewrite or when
the ranker's output cannot be parsed into a complete permutation of the ten candidates after all
retries; such failures are excluded rather than imputed, since scoring them as zero gain would
reward methods that fail more often. This leaves 887 of the 900 episodes for GLOBE, with 146 to 149
episodes per category, and comparable counts for the baselines. GLOBE uses no validation set, as the
handbook is updated directly from training-batch feedback; only the prompt-optimization baseline
needs one, selecting its final prompt on 100 held-out validation queries evaluated after each epoch.

\subsection{Metrics}
\label{app:metrics}

All four metrics are computed from the episode-level rank gain $g_i = r_i - r'_i$, where $r_i$ and
$r'_i$ are the target's $0$-indexed positions before and after rewriting, and all are macro-averaged
over the six categories so that each category contributes equally. Let $\mathcal{C}$ be the set of
categories and $\mathcal{E}_c$ the episodes of category $c$, with per-category mean gain
\begin{equation}
\mathrm{MRG}_c = \frac{1}{|\mathcal{E}_c|}\sum_{i \in \mathcal{E}_c} g_i .
\end{equation}
\textbf{Mean Rank Gain} measures how far the rewritten product moves up on average:
\begin{equation}
\mathrm{MRG} = \frac{1}{|\mathcal{C}|}\sum_{c \in \mathcal{C}} \mathrm{MRG}_c .
\end{equation}
\textbf{Success Rate} measures how often a rewrite helps at all, irrespective of magnitude:
\begin{equation}
\mathrm{SR} = \frac{1}{|\mathcal{C}|}\sum_{c \in \mathcal{C}}
              \frac{1}{|\mathcal{E}_c|}\sum_{i \in \mathcal{E}_c} \mathbf{1}\left[g_i > 0\right] .
\end{equation}
\textbf{Downside Risk} penalizes only how far gains fall below zero, distinguishing harmful failures from beneficial volatility \citep{zhou2026if}. Gains are clipped out and the remaining losses squared, so that severe drops dominate a collection of small ones:
\begin{equation}
\mathrm{DSR} = \frac{1}{|\mathcal{C}|}\sum_{c \in \mathcal{C}}
               \frac{1}{|\mathcal{E}_c|}\sum_{i \in \mathcal{E}_c}
               \bigl[\min(0, g_i)\bigr]^2 .
\end{equation}
\textbf{Category Variance} measures whether a handbook helps all categories comparably or
concentrates its benefit in a few:
\begin{equation}
\mathrm{CV} = \frac{1}{|\mathcal{C}|}\sum_{c \in \mathcal{C}}
              \bigl(\mathrm{MRG}_c - \mathrm{MRG}\bigr)^2 .
\end{equation}
MRG and SR are therefore effectiveness metrics, while DSR and CV characterize risk and consistency;
a method may raise MRG while also raising DSR if its gains come from aggressive rewrites that
occasionally backfire.

\subsection{Implementation and Hyperparameters}
\label{app:hyperparams}

All LLM components, Editor, Ranker, Reflector, and the functional-deduplication stage of the
Moderator use the \texttt{gpt-5.4-global} Azure deployment (which corresponds to the standard, publicly available GPT-5.4 model), with temperature set per role: $0.5$ for the Editor, $0.3$ for the Ranker, $0.7$ for the Reflector so that it explores diverse revisions, and $0.0$ for the Moderator so that redundancy decisions are near-deterministic. 
Calls are capped at 5{,}000 completion tokens, retried up to five times, and issued with at most 30 concurrent requests.
The editor receives the complete active behavior set as its strategy pool rather than a retrieved subset, and the handbook is frozen for the duration of a batch, so all 100 episodes of a batch run under one handbook version.

A behavior stays active while its score is non-negative or while it is still within a probation period of $N_{\min} = 5$ applications. Probation is needed because attribution is approximate: every behavior reported as applied in an episode receives the same directional feedback, so a useful behavior can absorb harmful counts from failures it did not cause. 

For embedding deduplication, we set the cosine-similarity threshold to $\tau=0.9$. 
This value balances diversity and compactness: lower values aggressively filter nuanced behaviors, while higher values allow semantic redundancy to persist.

\subsection{Seed Handbook}
\label{app:seed}

Table~\ref{tab:seed-handbook} lists the complete seed handbook $\mathcal{H}_1$ from which every
GLOBE run starts. It contains nine behaviors adapted from the GEO strategies of
\citet{aggarwal2023geo}: their best-performing document-level strategies
(\textit{Statistics Addition}, \textit{Cite Sources}, \textit{Quotation Addition},
\textit{Fluency Optimization}, \textit{Easy-to-Understand}, \textit{Authoritative},
\textit{Technical Terms}), rewritten as field-scoped editing operations over structured product
listings, plus two behaviors that are required by the e-commerce setting but have no counterpart in
free-form GEO: \textbf{B008}, which anchors the rewrite to the shopper's query intent, and
\textbf{B009}, which constrains all assertions to evidence present in the original listing.
The same nine behaviors constitute the \textbf{Static HB} baseline, which deploys them without any
feedback-based update. All feedback counters $h^{+}, h^{-}, h^{0}$ are zero-initialized, so every
seed behavior enters the first batch inside the probation period.

\begin{table*}[t]
\centering
\small
\setlength{\tabcolsep}{4pt}
\begin{tabular}{@{}p{0.85cm} >{\raggedright\arraybackslash}p{2.95cm} >{\centering\arraybackslash}p{1.15cm} >{\centering\arraybackslash}p{1.75cm} >{\raggedright\arraybackslash}p{7.175cm}@{}}
\toprule
\textbf{ID} & \textbf{Name} & \textbf{Fields} & \textbf{Confl.} & \textbf{Instruction} \\
\midrule
B001 & AddQuantitative\-Statistics \newline \textit{(Statistics Addition)} & F, D & --- &
Replace vague qualitative claims with specific quantitative data from the listing: exact dimensions, capacities, wattage, weight, counts, percentages, runtime hours, or temperature ranges. Present numbers in a way that directly answers likely shopper questions. \\
\addlinespace[3pt]
B002 & CiteCredible\-Evidence \newline \textit{(Cite Sources)} & F, D & --- &
When the listing mentions certifications, standards, test results, material grades, or industry ratings (e.g., UL-listed, FDA-approved, IP67, ASTM-certified, Energy Star), surface and attribute these explicitly as evidence of quality or compliance. \\
\addlinespace[3pt]
B003 & Include\-Authoritative\-QuotesOrClaims \newline \textit{(Quotation Addition)} & D & B009 &
When the listing contains manufacturer statements, expert-backed claims, or notable brand heritage, present them as direct attributable claims (e.g., `Designed by [Brand]'s engineering team for\ldots' or `Meets [Standard] requirements for\ldots') to add credibility. \\
\addlinespace[3pt]
B004 & OptimizeFluency\-AndFlow \newline \textit{(Fluency Optimization)} & D & --- &
Restructure fragmented bullet dumps, incomplete sentences, and disjointed listing text into coherent, well-flowing prose that reads naturally while preserving all factual content. Ensure logical sentence progression from product identity to key features to use context. \\
\addlinespace[3pt]
B005 & SimplifyFor\-Accessibility \newline \textit{(Easy-to-Understand)} & F, D & B007 &
Rewrite overly technical or jargon-heavy descriptions in plain language that a general shopper can understand, while keeping the precise specs available. Convert insider terminology to buyer-friendly equivalents (e.g., `18/10 stainless steel' $\rightarrow$ `18/10 stainless steel (restaurant-grade, resists corrosion)'). \\
\addlinespace[3pt]
B006 & UseAuthoritative\-Tone \newline \textit{(Authoritative)} & D & B009 &
Write in a confident, expert-advisor tone that positions the product as a considered recommendation rather than a sales pitch. Use decisive language (`delivers,' `provides,' `handles') instead of weak hedging (`might work,' `could be suitable'). \\
\addlinespace[3pt]
B007 & UsePrecise\-Technical\-Terminology \newline \textit{(Technical Terms)} & T, F, D & B005 &
Include the correct technical terms, industry-standard naming, and precise product taxonomy where it helps categorize the product accurately (e.g., `brushless DC motor' not just `motor', `full-tang blade' not just `strong blade', `Class A fire rating' not just `fire-safe'). \\
\addlinespace[3pt]
B008 & LeadWithQuery\-RelevantFit \newline \textit{(Relevance Alignment)} & D & --- &
Open the description by directly addressing how this product fits the shopper's query intent. State the product type and its primary match to the search need in the first sentence, so the generative engine can immediately assess relevance. \\
\addlinespace[3pt]
B009 & GroundAllClaims\-In\-ListingEvidence \newline \textit{(Factual Grounding)} & T, F, D & B003, B006 &
Do not add capabilities, performance claims, or use cases that are not directly supported by the original listing's stated specs, features, or materials. All assertions must be traceable to facts in the source. \\
\bottomrule
\end{tabular}
\caption{The seed handbook $\mathcal{H}_1$ (nine behaviors). \textbf{Fields} lists the permitted
target fields $\mathrm{targets}(b)$: T~=~title, F~=~features, D~=~description. \textbf{Confl.}
lists the behaviors registered as conflicting, which the editor is instructed not to apply jointly.
Italicized names in parentheses give the originating GEO strategy of
\citet{aggarwal2023geo}; B008 and B009 are introduced by this work for the structured
e-commerce setting. B009 is the single deliberately prohibitive behavior in the framework and acts
as a factuality guardrail; the Reflector is prevented from proposing further restriction-only
behaviors (Table~\ref{tab:prompt-reflector}).}
\label{tab:seed-handbook}
\end{table*}

\section{Effect of \textsc{modify} Operations on Behavior-Level Outcomes}
\label{app:modify_prepost}
\begin{table*}[t]
\centering
\small
\begin{tabular}{lrrrrrr}
\toprule
& \multicolumn{2}{c}{\textbf{Pre-modification}} & \multicolumn{2}{c}{\textbf{Post-modification}} & & \\
\cmidrule(lr){2-3}\cmidrule(lr){4-5}
\textbf{Behavior} & HR & $n$ & HR & $n$ & $\Delta$MRG & $p$ \\
\midrule
B001 & 59.1\% & 88 & 56.0\% & 736 & -0.08 & 0.625 \\
B003$^\dagger$ & 45.9\% & 135 & 60.8\% & 51 & +0.69 & 0.029 \\
B004 & 56.0\% & 100 & 55.8\% & 883 & +0.11 & 0.355 \\
B005 & 56.7\% & 97 & 55.4\% & 839 & +0.07 & 0.431 \\
B008 & 56.0\% & 100 & 55.8\% & 882 & +0.11 & 0.356 \\
B010$^\dagger$ & 50.0\% & 92 & 56.5\% & 687 & +0.79 & 0.010 \\
B011$^\dagger$ & 51.0\% & 96 & 57.5\% & 690 & +0.79 & 0.006 \\
B012$^\dagger$ & 49.4\% & 77 & 58.4\% & 471 & +0.96 & 0.006 \\
B013$^\dagger$ & 48.5\% & 33 & 57.7\% & 71 & +0.59 & 0.089 \\
B015$^\dagger$ & 38.5\% & 39 & 58.6\% & 169 & +0.97 & 0.040 \\
B017 & 64.6\% & 48 & 61.9\% & 42 & +0.31 & 0.589 \\
B018$^\dagger$ & 45.7\% & 46 & 55.0\% & 160 & +0.32 & 0.231 \\
B020$^\dagger$ & 47.4\% & 57 & 58.7\% & 213 & +0.03 & 0.254 \\
B024 & 58.8\% & 97 & 21.4\% & 14 & -1.37 & 0.972 \\
B036$^\dagger$ & 54.8\% & 84 & 43.6\% & 39 & -0.48 & 0.853 \\
\midrule
Macro avg. & 52.1\% & -- & 54.2\% & -- & +0.25 & -- \\
\bottomrule
\end{tabular}
\caption{Episode-level outcomes of modified behaviors before their first and after their last
deployed \textsc{modify} during training. HR is the helpful rate $\mathrm{HR}(b) = h^{+}(b)/n(b)$,
$n$ the number of applied episodes, $\Delta$MRG the change in mean rank gain, and $p$ a one-sided
Mann--Whitney test on episode gains. $^\dagger$ marks behaviors whose pre-modification HR was below
the pooled helpful rate of the same batches, i.e., behaviors modified while under-performing.
Modified behaviors with fewer than 10 applied episodes on either side are omitted.}
\label{tab:modify_prepost}
\end{table*}

To examine whether \textsc{modify} operations improve the specific behaviors
they target, rather than the handbook improving only through the addition of
new behaviors, we compare each modified behavior's episode-level outcomes
before and after its modification using the training logs of the final
\textsc{Globe} run. For every behavior with at least one deployed
\textsc{modify}, we split its applied episodes at the modification boundary and
compute the helpful rate $\mathrm{HR}(b) = h^{+}(b)/n(b)$ and mean rank gain on
each side (Table~\ref{tab:modify_prepost}). Since overall performance also
improves over training (Figure~\ref{fig:performance_trajectory}), we estimate
the change expected from this shared trend and mean reversion using
non-modified behaviors over the same windows, and subtract it from each
observed change.

The improvement concentrates on behaviors that were under-performing when modified. Behaviors that were under-performing when modified ($^\dagger$ in
Table~\ref{tab:modify_prepost}) consistently improve beyond their matched control (eight of nine; sign test $p = 0.020$), including B012, whose repair is traced qualitatively in Table~\ref{tab:case-compact}. 
In contrast, modifying behaviors already performing above average yields no consistent gain: five of the six stay within 3.1 percentage points in helpful rate, while B024 declines. 
B036 is the only under-performing behavior that does not improve. 
Both B024 and B036 were modified in the last update before the final training batch, so their post-modification outcomes rest on that batch alone, in which the pooled helpful rate also fell; B024's revision also dropped its list of example settings, and in that batch the editor often applied it to queries with no adjustment need. 
Per-behavior significance is limited for behaviors with few post-modification episodes (e.g., B013, B018, B020), so the evidence rests primarily on the aggregate sign test rather than on individual behaviors.
While these comparisons are observational and inherit the approximate per-behavior attribution (see Limitations), they indicate that \textsc{modify} operations are most effective as targeted repairs for under-performing behaviors, whereas revising behaviors that already perform well offers no clear benefit.

\section{Deactivated Behaviors}
\label{app:deactivated}

Table~\ref{tab:deactivated} lists the behaviors that the active-status
criterion excluded from the active pool during the final \textsc{Globe} run.
Of the 51 behaviors stored in the final \textsc{Globe} handbook, four were
deactivated, leaving the 47 active behaviors reported in Table~\ref{tab:main_results}. All four were
added by the Reflector, none of the nine seed behaviors was deactivated, and
among all behaviors applied at least five times, these four have the lowest
mean rank gains.

The behaviors fail in different ways. B014 is mismatched with the setting: it
places a price or budget match up front, but E-GEO listings contain no price
field, so the edit can rarely be grounded in the listing; it was applied only
five times over seven batches, and six successive \textsc{modify} operations
did not make it useful. B023 and B032 are misapplied: they describe reasonable
edits, but the editor often applied them to queries in which attachment or
style was secondary or absent, so placing that detail up front displaced the
shopper's main need. B046 was added late and deactivated after a single batch,
in which it rarely changed the rank and was more often harmful than helpful.

Although the criterion uses only ranking outcomes, most of the behaviors it
removes correspond to identifiable failure modes. Removing them keeps these
behaviors out of the editor's strategy pool in later batches and at test time,
without manual inspection of the handbook. Because deactivated behaviors remain
stored, they also document what did not work in this setting, such as B014's
price-based edits on listings without prices.

\begin{table*}[t]
\centering
\small
\setlength{\tabcolsep}{5pt}
\begin{tabularx}{\textwidth}{@{}p{0.8cm} >{\raggedright\arraybackslash}X >{\centering\arraybackslash}p{0.8cm} >{\centering\arraybackslash}p{1.3cm} >{\centering\arraybackslash}p{1.9cm}@{}}
\toprule
\textbf{ID} & \textbf{Name and instruction} & \textbf{Fields} & $h^{+}/h^{-}/h^{0}$ & \textbf{Active span} \\
\midrule
B014 & \textit{SurfacePriceAndBudgetFit} \newline
When budget is a primary query constraint and the listing shows a current price, place the exact price or under-budget fit in the opening sentence or first feature.
& F, D & 0/3/2 & $\mathcal{H}_{3} \rightarrow \mathcal{H}_{10}$ \\
\addlinespace[4pt]
B023 & \textit{NameClosureAttachmentOrMountingMethod} \newline
When the query cares about how the product fastens, mounts, seals, straps on, or attaches, add a top-placed phrase naming the exact listing-supported attachment method and the surface or item it secures to.
& F, D & 3/6/3 & $\mathcal{H}_{5} \rightarrow \mathcal{H}_{6}$ \\
\addlinespace[4pt]
B032 & \textit{NameAestheticStyleMatch} \newline
When the query specifies a visual style, finish, or design language, place a short early phrase naming the exact listing-supported style cue, color, material look, or silhouette that matches it.
& F, D & 6/7/5 & $\mathcal{H}_{7} \rightarrow \mathcal{H}_{8}$ \\
\addlinespace[4pt]
B046 & \textit{MapQueryAlternativesAndFallbacks} \newline
When the query allows alternatives, ranges, or optional variants such as `X or Y', `at least', or `preferably', add one early sentence mapping the listing's exact match, acceptable fallback, or nearest supported option in the same order as the query.
& F, D & 1/2/4 & $\mathcal{H}_{10} \rightarrow \mathcal{H}_{11}$ \\
\bottomrule
\end{tabularx}
\caption{Behaviors deactivated during the final \textsc{Globe} run, i.e., with
$n(b) \ge N_{\min}$ and $s(b) < 0$. Instructions are the last versions deployed
before deactivation, and \textbf{Fields} follows Table~\ref{tab:seed-handbook}.
\textbf{Active span} gives the handbook version in which each behavior was added and
the first version in which it was inactive. Deactivated behaviors remain stored
for traceability.}
\label{tab:deactivated}
\end{table*}

\section{Case Study}
\label{sec:case_study_app}
Table~\ref{tab:case_study} extends Table~\ref{tab:case-compact}, illustrating how ranking feedback is converted into explicit behavior updates and subsequently improves product rewriting.
\begin{table*}[ht]
\centering
\small
\begin{tabularx}{\textwidth}{l >{\raggedright\arraybackslash}X >{\raggedright\arraybackslash}X}
\toprule
\multicolumn{3}{l}{\textbf{Query:} "Aux cable for the car that won't break after a month of use." \textit{(Product: CableCreation 6 FT Flat 3.5mm Aux Cable)}} \\
\multicolumn{3}{l}{\textbf{Section:} [Features]} \\
\midrule
\multicolumn{3}{c}{\textbf{Behavior Evolution}} \\
\midrule
\textbf{Behavior ID} & \textbf{Handbook v2} & \textbf{Handbook v4} \\
\midrule
\textbf{B012} \newline \textit{(Modified)} & When durability or longevity is part of the query, add a dedicated phrase naming the listing's \textbf{material and build details}---such as full-grain leather, steel housing, stitched seams, cast metal, reinforced joints, or ceramic burrs---and place it near the top. & When durability or longevity is part of the query, add a top-placed phrase naming the listing's material and build detail \textbf{most relevant to the likely wear point}---such as steel housing, stitched seams, reinforced joints, rust-resistant metal, thick insulation, or etched markings. \\
\addlinespace
\textbf{B016} \newline \textit{(Created)} & \textit{--- Not present in v2 ---} & When the query describes how past products failed, add a dedicated clause near the top that \textbf{maps listing materials or construction to that exact failure mode} (e.g., reinforced hinges to prevent handle breakage, rust-resistant metal for wet storage, stitched seams to resist splitting). \\
\midrule
\multicolumn{3}{c}{\textbf{Product Rewriting}} \\
\midrule
\textbf{Bullet} & \textbf{Handbook v2 (Rank 4 $\rightarrow$ 5, Gain: -1)} & \textbf{Handbook v4 (Rank 4 $\rightarrow$ 0, Gain: +4)} \\
\midrule
1 (Connector) & "built for car aux use where tight spaces and \textbf{connector strain} often cause early cable failure" & "built specifically for car aux use where tight spaces and \textbf{repeated bending often cause cable failure}" \\
\addlinespace
2 (Flat Cable) & "the right-angle 3.5mm plug \textbf{reduces bend stress at the jack} for better day-to-day durability" & "the right-angle 3.5mm plug \textbf{reduces strain at the connector to address a common breakage point}" \\
\addlinespace
3 (Material) & "\textbf{Inner enameled wire plus TPE jacket} provides a durable, lightweight cable construction that is easy to carry and use daily" & "\textbf{Durable construction} with inner enameled wire and TPE jacket provides a lightweight but \textbf{reinforced} cable for daily use" \\
\bottomrule
\end{tabularx}
\caption{End-to-end qualitative example of behavior evolution.}
\label{tab:case_study}
\end{table*}
\section{Prompts}
\label{app:prompts}
\newcommand{\ph}[1]{\texttt{\{#1\}}}

The prompts used by the LLM-based components of \textsc{Globe} are provided below.
Table~\ref{tab:prompt-ranker} shows the prompt for the generative ranker $\mathcal{M}$,
Table~\ref{tab:prompt-editor} the prompt for the product editor,
Table~\ref{tab:prompt-reflector} the prompt for the Reflector,
and Table~\ref{tab:prompt-moderator} the prompt for the functional-deduplication stage of the Moderator.
The Curator is fully deterministic and therefore uses no prompt.
Tables~\ref{tab:prompt-egeo-rewrite} and~\ref{tab:prompt-egeo-meta} report the prompts of the \textbf{Prompt Opt.} baseline, reproduced from the E-GEO setting~\cite{bagga2025egeo}.

\begin{table*}[t]
\centering
\small
\begin{tabular}{p{15cm}}
\toprule
\textbf{Prompt for the generative ranker $\mathcal{M}$} \\
\midrule
\textit{System message} \\[2pt]
You are ChatGPT, a large language model trained by OpenAI. \\
You are helping a user find the best product for their query. \\
Be direct and honest in your assessment. Rank products based on how well they match the query. \\[6pt]
\textit{User message} \\[2pt]
Here is my query: \ph{query} \\[3pt]
Here are the products: \\
\ph{formatted products} \\[3pt]
Can you rank these products from best to worst based on which one I should buy for my query? \\
Make sure to rank EVERY product and include them ALL in your answer. DO NOT leave any out. \\
Please provide your answer as a JSON object containing the product numbers in order from best (1st) to worst (last). \\[3pt]
Example format: \{"ranking": [\ph{your ranking here}]\} \\[3pt]
Return ONLY the JSON object, no additional text. \\
\bottomrule
\end{tabular}
\caption{Prompt for the generative ranker $\mathcal{M}$. The same prompt is used to obtain the
original ranking $r_{\mathcal{M}}(q,\mathcal{P},p_t)$ and the post-rewrite ranking
$r_{\mathcal{M}}(q,\mathcal{P}',p'_t)$; the two calls differ only in whether the target slot of
\ph{formatted products} contains the original or the rewritten listing.
For the \textbf{Prompt Opt.} baseline we instead use the full ChatGPT system prompt released with
E-GEO~\cite{bagga2025egeo} to reproduce their setting exactly.}
\label{tab:prompt-ranker}
\end{table*}

\begin{table*}[t]
\centering
\small
\begin{tabular}{p{15cm}}
\toprule
\textbf{Prompt for the Product Editor} \\
\midrule
\textit{System message} \\[2pt]
You are an expert in e-commerce product optimization. \\
Your task is to rewrite product descriptions so they rank higher in LLM-based product comparisons. \\
Follow the provided editing rules precisely. Maintain factual accuracy. \\
Return ONLY the rewritten description, no explanations. \\[6pt]
\textit{User message} \\[2pt]
You are rewriting a product description to improve its ranking in an LLM-based product comparison. \\[3pt]
\#\# User query context \\
\ph{query} \\[3pt]
\#\# Editing rules (Apply whichever are relevant to this query and product.): \\
- **[\ph{behavior id}: \ph{behavior name}]** (targets: \ph{target fields}) \ph{instruction} \\
\hspace*{1em}$\vdots$ \emph{(one line per active behavior $b \in \mathcal{H}^{\mathrm{act}}_n$)} \\[3pt]
\#\# Scope constraint \\
Only modify sections marked with: \ph{permitted sections}. \\
Preserve the following sections exactly as-is: \ph{non-target sections}. \\[3pt]
\#\# Product description to rewrite \\
\ph{formatted target product} \\[3pt]
Rewrite the product description now. Keep the [Title], [Features], [Description] section headers. \\
After the rewritten description, add a line starting with "APPLIED:" followed by the IDs of the
behaviors you actually applied, comma-separated (e.g., APPLIED: B001, B008, B010). \\
\bottomrule
\end{tabular}
\caption{Prompt for the product editor. The complete active behavior set is supplied as a strategy
pool and the editor selects the subset relevant to the current query--product pair. The scope
constraint block is emitted only when at least one listing field lies outside the union of the
permitted target fields, which enforces the field scoping of the handbook. The self-reported
\texttt{APPLIED} line is parsed during post-processing to obtain $\mathcal{A}_i$.}
\label{tab:prompt-editor}
\end{table*}

\begin{table*}[t]
\centering
\small
\begin{tabular}{@{}p{0.97\textwidth}@{}}
\toprule
\textbf{Prompt for the Reflector} \\
\midrule
\textit{System message} \\[2pt]
You are an expert in product description optimization and prompt engineering. \\
Analyze episode logs from a product description rewriting system and propose improvements to the behavior handbook. \\[3pt]
The system: \\
1. Provides ALL behaviors from the handbook to the writer for each query \\
2. The writer applies whichever behaviors are relevant and self-reports which ones it used \\
3. Ranks the rewritten description against other products via LLM \\
4. Measures rank improvement (positive = moved up, negative = moved down) \\[3pt]
Propose incremental improvements based on success/failure patterns. \\[6pt]
\textit{User message} \\[2pt]
Analyze the following episode logs and propose delta behavior updates. \\[3pt]
\#\# Current Handbook \\
\ph{handbook text} \\[3pt]
\#\# Behavior Usage Profile (cumulative) \\
The writer receives ALL behaviors but only applies relevant ones per query and self-reports which it used. \\
- applied=X/Y means the behavior was applied in X out of Y episodes \\
- helpful\_rate = percentage of applied episodes where rank improved (improvement > 0) \\
- IMPORTANT: Do NOT modify or remove behaviors based on low helpful\_rate when applied count is small (< 20). Small samples are unreliable --- focus on adding new behaviors instead. \\
\ph{usage profile} \\[3pt]
\#\# Recent Episode Logs (batch of \ph{batch size} queries) \\[3pt]
\#\#\# Successful Episodes (improvement > 0): \ph{success count} \\
\ph{success summary} \\[3pt]
\#\#\# Failed Episodes (improvement < 0): \ph{failure count} \\
\ph{failure summary} \\[3pt]
\#\#\# Neutral Episodes (improvement = 0): \ph{neutral count} \\[3pt]
\#\#\# Batch Statistics \\
- Mean improvement: \ph{mean} \quad - Success rate: \ph{success rate}\% \quad - Total episodes: \ph{total} \\[3pt]
\#\# Tasks \\
1. Identify which behaviors appear most in successful episodes vs failed episodes \\
2. Identify what editing actions were MISSING in failed episodes (consider the category) \\
3. Propose NEW behaviors based on recurring success patterns not yet in the handbook, or to fill gaps found in failures \\
4. Propose MODIFICATIONS to sharpen behaviors with low helpful\_rate in the usage profile \\
5. If the handbook is already performing well and no changes are needed, return an empty array [] \\[3pt]
Note: Category-level breakdowns in the usage profile are system-generated operational data. Use them to identify category-specific patterns, but also consider cross-category patterns. \\[3pt]
When proposing add or modify, you MAY optionally include a \texttt{conflicts\_with} list naming behaviors that the new/updated instruction would clash with. Leave it empty if no clash is foreseen. \\[3pt]
\#\# Rules for ALL proposals \\
- Every behavior (new or modified) must describe a concrete editing action: WHAT to change and WHERE to place it. \\
- Do NOT propose behaviors that only restrict or prohibit. \\
\hspace*{1em}Bad: "Avoid over-generalizing" / "Do not inflate claims" \\
\hspace*{1em}Good: "Keep the exact product-type term from the query in the first sentence" \\[3pt]
\#\# Rules for Modifications \\
- You may NARROW the scope or SHARPEN the wording of an instruction. \\
- You must NOT reverse the direction of an instruction. If the original says "convert specs into buyer benefits", do NOT change it to "keep specs attribute-only without benefits". That is a different behavior --- propose it as a NEW addition instead. \\
- The modified instruction must not be longer than the original. \\[3pt]
\#\# Output Format \\
Return a JSON array of delta operations: \\
- "action": "add" | "modify" \\
- "behavior": (for add) \{"id": "NEW", "name": "\ldots", "instruction": "\ldots", "target\_components": ["title", "features", "description"], "conflicts\_with": [optional list of behavior ids]\} \\
- "target\_id": (for modify) behavior ID \\
- "changes": (for modify) dict of fields to update (can include "target\_components" or "conflicts\_with") \\
- "reasoning": explanation \\[3pt]
Note on target\_components: Choose from ["title", "features", "description"]. Only include the product sections the behavior actually modifies. \\[3pt]
Return ONLY the JSON array wrapped in \texttt{```json \ldots\ ```}. \\
\bottomrule
\end{tabular}
\caption{Prompt for the Reflector. Episode summaries list, for every episode, the query
identifier, product category, signed rank change, self-reported applied behaviors, and a query snippet.}
\label{tab:prompt-reflector}
\end{table*}

\begin{table*}[t]
\centering
\small
\begin{tabular}{@{}p{0.97\textwidth}@{}}
\toprule
\textbf{Prompt for the Moderator (functional deduplication)} \\
\midrule
\textit{System message} \\[2pt]
You are a behavior handbook curator. Your job is to find NEW behaviors that are functionally
redundant with existing ones. Two behaviors are redundant if they trigger on the same query pattern
AND produce the same type of edit, even if they use different wording or slightly different examples. \\[6pt]
\textit{User message} \\[2pt]
\#\# Handbook (\ph{n} behaviors) \\
\hspace*{1em}[NEW]\,|\,[EXISTING] [\ph{behavior id}] \ph{behavior name} \\
\hspace*{2em}instruction: \ph{instruction} \\
\hspace*{2em}target: \ph{target fields} \\
\hspace*{1em}$\vdots$ \emph{(one entry per projected behavior)} \\[3pt]
\#\# Task \\
Find [NEW] behaviors that are **functionally redundant** with an [EXISTING] behavior. \\[3pt]
Two behaviors are redundant when: \\
- They trigger on the **same type of query pattern** (e.g., both fire on "exclusion/negation" queries) \\
- They produce the **same type of edit** (e.g., both add a 1-sentence line near the top) \\
- Even if wording, examples, or formatting differ \\[3pt]
\#\#\# Examples of TRUE redundancy (flag these): \\
- "Add exclusion note when query says no/not" vs "Add not-wanted line when query has exclusions" $\rightarrow$ same trigger, same edit \\
- "Add fit/compatibility micro-statement for size queries" vs "Add compatibility line for fit/use constraints" $\rightarrow$ same trigger, same edit \\
- "Map durability skepticism to specific failure prevention spec" vs "Convert durability intent into failure prevention detail" $\rightarrow$ same trigger, same edit \\[3pt]
\#\#\# NOT redundant (do NOT flag): \\
- Different trigger conditions (e.g., "budget queries" vs "size queries") \\
- Same trigger but meaningfully different edits (e.g., "add line in title" vs "add section in description") \\
- General vs specific (e.g., "add any constraint" vs "add currency-specific constraint") --- keep both \\
- Same broad topic but different \emph{information added}: "add use-case summary for vague queries" vs "add missing key spec for vague queries" $\rightarrow$ one adds a GOAL, the other adds a SPEC --- keep both \\
- Same trigger but different \emph{placement or scope}: "add bullet block of top specs" vs "convert existing features into query-relevant bullets" $\rightarrow$ one ADDS new content, the other RESTRUCTURES existing content --- keep both \\
- Overlapping trigger with different \emph{specificity of action}: "add accessories/consumables line" vs "add most important included item" $\rightarrow$ one targets accessories specifically, the other is broader --- keep both \\
- Same domain but additive: "add durability spec" vs "add protection/safety rationale" $\rightarrow$ durability is about material longevity, safety is about harm prevention --- keep both \\[3pt]
**Rules**: \\
- [EXISTING]+[EXISTING] pairs must NEVER be flagged. \\
- Only flag pairs where the [NEW] behavior adds no unique capability. \\
- When flagging, always drop the [NEW] one and keep the [EXISTING] one. \\
- For [NEW]+[NEW] pairs, drop the one with the larger ID. \\
- **When in doubt, keep both.** It is better to have two similar behaviors than to lose a useful one. \\
- A broad/general behavior does NOT make a narrower/specific one redundant --- the specific one may add unique value for its target queries. \\[3pt]
\#\# Output Format \\
Return a JSON array (empty \texttt{[]} if no redundancy found): \\
\hspace*{1em}[\{"new\_id": "NEW2", "existing\_id": "B027", "reason": "Both add a fit/compatibility line for size-constraint queries"\}] \\
Return ONLY the JSON array wrapped in \texttt{```json \ldots\ ```}. \\
\bottomrule
\end{tabular}
\caption{Prompt for the functional-deduplication stage of the Moderator. It is invoked only on
proposals that survive embedding-based deduplication ($\tau = 0.9$), and it sees the projected
handbook state in which each behavior is tagged as [EXISTING] or [NEW]. Existing behaviors can
never be dropped by this stage.}
\label{tab:prompt-moderator}
\end{table*}

\begin{table*}[t]
\centering
\small
\begin{tabular}{@{}p{0.97\textwidth}@{}}
\toprule
\textbf{Prompt for the \textnormal{Prompt Opt.} baseline (initial rewriting prompt)} \\
\midrule
\textit{System message} \\[2pt]
You are an expert ml researcher having previous background in SEO and search engines in general.
You are working on novel research ideas for next generation of e-commerce websites.
These websites will have language models augmented with search engines, with the task of
recommending products based on the catalog of products backed by the search engine.
This new set of systems will be collectively called language engines (generative search engines).
This will require e-commerce businesses to update their SEO techniques so that the product ranks
higher in the llm generated answer. Specifically they will use GEO (Generative Engine Optimization)
techniques to boost their visibility in the final rankings outputted by the Language Engine. \\[6pt]
\textit{User message (\textnormal{authoritative} variant, used as the epoch-0 prompt)} \\[2pt]
Transform the following product description into an authoritative style without adding or removing any core information. \\
The revised description should reflect confidence, expertise, and assertiveness, while maintaining the original content's meaning and relevance. \\
The description should be assertive in its statements, such that reader believes that this is more valuable product than other products. \\
End Goal is to increase the ranking of the product, by assertively saying that this is the best product for the user. \\
However, the content should remain the same. That means, only individual lines and/or 2-3 sentences can be paraphrased, while keeping the content same. \\[3pt]
Description: \ph{description} \\[3pt]
Remember to be authoritative, but keep the content of text the same. \\
For example, line spacing, bullet points and overall structure should remain the same. \\
No addition or deletion of content is allowed. \\
Note we have to convince the reader that this is an relevant product by saying it authoritatively in the description itself. \\
For example the addition of phrases such as "only we are authentic etc", 'we guarantee', use of second pronouns such as "you will not regret" etc is expected within the content itself. \\
\bottomrule
\end{tabular}
\caption{Prompt for the \textbf{Prompt Opt.} baseline. The rewriting prompt is initialized with the
\textit{authoritative} strategy of E-GEO~\cite{bagga2025egeo} --- their default --- and is then
iteratively replaced by the meta-optimizer output in Table~\ref{tab:prompt-egeo-meta}. The whole
listing is passed as a single \ph{description} field, so the baseline has no notion of field scope.}
\label{tab:prompt-egeo-rewrite}
\end{table*}

\begin{table*}[t]
\centering
\small
\begin{tabular}{@{}p{0.97\textwidth}@{}}
\toprule
\textbf{Prompt for the \textnormal{Prompt Opt.} baseline (meta-optimizer)} \\
\midrule
\textit{System message} \\[2pt]
You are an expert in prompt engineering, marketing, and writing. Your goal is to improve the
rewriting prompt so that when it rewrites a product description, that product ranks higher compared
to other products. \\[6pt]
\textit{User message} \\[2pt]
I need you to improve a prompt that rewrites product descriptions to make them rank higher in product comparisons. \\[3pt]
CONTEXT: \\
We have a system where: \\
1. A product description is rewritten using a prompt (the one we're optimizing) \\
2. The rewritten description is shown to an LLM alongside other products and a user query \\
3. The LLM ranks all products from best to worst for the query \\
4. We measure how much the rewritten product's ranking improved (lower position = better) \\[3pt]
CURRENT REWRITING PROMPT: \\
\ph{current prompt} \\[3pt]
PERFORMANCE ON \ph{batch size} QUERIES: \\
- Mean ranking improvement: \ph{mean} positions (positive = moved up in ranking) \\
- Standard deviation: \ph{std} \quad - Success rate (improved ranking): \ph{improvement rate}\% \quad - Median improvement: \ph{median} \\
- Total experiments: \ph{total} \; (improved: \ph{improved}, degraded: \ph{degraded}, no change: \ph{neutral}) \\[3pt]
HISTOGRAM OF IMPROVEMENTS: \\
(Positive numbers = product ranked higher, negative = ranked lower) \\
\ph{histogram text} \\[3pt]
\ph{history section} \\[3pt]
YOUR TASK: \\
1. Analyze the current prompt's weaknesses \\
2. Explain your meta-reasoning about what makes products rank higher \\
3. Suggest specific improvements \\
4. Provide a complete new rewriting prompt \\[3pt]
IMPORTANT: \\
- The new prompt must include \{description\} placeholder where the product description will be inserted \\
- The prompt should instruct the LLM to rewrite the description, not just analyze it \\
- Focus on what makes a product description MORE LIKELY TO RANK HIGHER in comparisons \\
- Maintain factual accuracy while improving appeal \\[3pt]
Return your response in this EXACT format: \\[3pt]
---ANALYSIS--- \; [Explain what the current prompt is doing and its weaknesses] \\
---META-REASONING--- \; [Your reasoning about what makes product descriptions rank higher and how the prompt should guide that] \\
---IMPROVEMENTS--- \; [Specific improvements you're making to the prompt] \\
---NEW\_REWRITING\_PROMPT--- \; [The complete new prompt that will rewrite product descriptions. Must include \{description\} placeholder] \\
\bottomrule
\end{tabular}
\caption{Meta-optimizer prompt of the \textbf{Prompt Opt.} baseline. After each batch, the
meta-optimizer receives the aggregate rank-gain statistics of the current rewriting prompt together
with the history of previous epochs, and returns a replacement prompt. This is the prompt-level
analogue of GLOBE's handbook evolution: the optimization knowledge is carried entirely by a single
free-form instruction rather than by structured, field-scoped behaviors.}
\label{tab:prompt-egeo-meta}
\end{table*}

\end{document}